\documentclass{article} % For LaTeX2e

\usepackage{arxiv,times} % PG- use the gram style file here for the conference
\usepackage{tikz}
\usetikzlibrary{positioning, shapes.geometric, arrows.meta, calc, fit, backgrounds}
\usepackage{amsmath}
\usepackage[most]{tcolorbox}
\usepackage{mdframed}

\maintrue % PG --- set this for main proceedings, else it uses defaults for the header and footer to say tiny paper track
\usepackage{enumitem}

\usepackage{hyperref}
\usepackage{url}
\usepackage{graphicx}
\usepackage{placeins}
\usepackage{wrapfig}
\usepackage{amssymb}
\usepackage{etoc}

\title{Mutual information and task-relevant latent dimensionality}

\usepackage{authblk}

\author[1, *]{Paarth Gulati}
\author[2, *]{Eslam Abdelaleem}
\author[2]{Audrey Sederberg}
\author[1,3]{Ilya Nemenman}

\affil[1]{Department of Physics, Initiative for Theory and Modeling of Living Systems, Emory University}
\affil[2]{School of Physics, School of Psychology, Georgia Institute of Technology}
\affil[3]{Department of Biology, Emory University}

\renewcommand{\headrulewidth}{0pt}
\iclrfinalcopy % Uncomment for camera-ready version, but NOT for submission.
\begin{document}

\etocdepthtag.toc{mtoc} % Tag main text sections as 'mtoc'

\maketitle
\begingroup
\renewcommand\thefootnote{*}
\footnotetext{P.G. and E.A. contributed equally to this work.}
\endgroup
\begin{abstract}
Estimating the dimensionality of the latent representation needed for prediction---the task-relevant dimension---is a difficult, largely unsolved problem with broad scientific applications. We cast it as an Information Bottleneck question: what embedding bottleneck dimension is sufficient to compress predictor and predicted views while preserving their mutual information (MI). This repurposes neural MI estimators for dimensionality estimation. We show that standard neural estimators with separable/bilinear critics systematically inflate the inferred dimension, and we address this by introducing a hybrid critic that retains an explicit dimensional bottleneck while allowing flexible nonlinear cross-view interactions, thereby preserving the latent geometry. We further propose a one-shot protocol that reads off the effective dimension from a single over-parameterized hybrid model, without sweeping over bottleneck sizes. We validate the approach on synthetic problems with known task-relevant dimension. We extend the approach to intrinsic dimensionality by constructing paired views of a single dataset, enabling comparison with classical geometric dimension estimators. In noisy regimes where those estimators degrade, our approach remains reliable. Finally, we demonstrate the utility of the method on multiple physics datasets.
\end{abstract}

\section{Introduction}
% \footnotetext[1]{P.G. and E.A. contributed equally to this work.}

Before ``low-dimensional latent embeddings'' became a rallying cry of AI, they were already a basic aim of science: identify a low-dimensional \emph{state}---a small set of degrees of freedom constructed from observations---that suffices to predict the quantities of interest. The long road from Aristotelian to Newtonian mechanics illustrates that  determining the \emph{number} of such state variables---the relevant \emph{latent dimensionality}---can be hard, even before one argues about the right variables or the laws that relate them. In today’s high-throughput, AI-enabled scientific world, the (somewhat Wignerian) ``unreasonable effectiveness'' \citep{Wigner1960} of low-dimensional descriptions in modeling complex physical systems is reinforcing the view that such latent structure is often a property of the underlying systems rather than a modeling choice \citep{huh2024platonic,edamadaka2025universally}. The evidence for this spans fluid dynamics \citep{chen2022automated}, molecular structure and dynamics \citep{das2006low,tamura2022structural}, jammed materials \citep{cubuk2017structure,jin2021jamming}, and neural population activity \citep{gallego2017neural,semedo2019cortical,schneider2023learnable}. Yet estimates of the \emph{relevant} latent dimensionality often vary widely across methods, even within the same study.

This problem has two roots. First, for noisy scientific data an ``intrinsic'' dimensionality of the raw observations is neither unique nor especially useful: for example, the variables needed to predict a body’s future position are not the same as those needed to record its shape.   For scientific applications, one needs instead a \emph{task-relevant} dimension---the size of the minimal state  required to preserve the information relevant for a specified prediction problem.  While there are niche attempts to adapt  intrinsic dimension notions to such task-relevant  settings, there is no accepted general solution. Second, even if one asks only for an intrinsic dimension of the data distribution, estimation is notoriously fragile in the high-dimensional, undersampled regime common in science. Both classical nonlinear-dynamics estimators \citep{grassberger1983measuring} and modern neighbor-statistics methods such as Levina--Bickel \citep{levina2004maximum} and Two-NN \citep{facco2017estimating} (see also \citep{camastra2016intrinsic}) can return convincing but meaningless estimates  on limited, noisy data \citep{theiler1986spurious}. More fundamentally, for realistic dataset sizes these approaches often saturate at dimensions ``like 6 or 7'' \citep{eckmann1992fundamental}---a vintage
ancestor of a modern meme.

In this paper, we use mutual information (MI) to formalize task relevance. We  leverage recent advances in neural MI estimation \citep{Belghazi2018MutualEstimation,oord2018representation,poole2019variational,song2019understanding} to estimate the dimensionality. Given paired views $(X,Y)$---a predictor and the quantity to be predicted---we seek the smallest bottleneck dimension $k_z$ for which compressed representations $Z_X=f(X)\in\mathbb{R}^{k_z}$ and $Z_Y=g(Y)\in\mathbb{R}^{k_z}$ preserve the shared information, i.e., $I(Z_X;Z_Y)\approx I(X;Y)$. This symmetric information bottleneck (SIB) viewpoint \citep{friedman2013multivariate,abdelaleem2025deep} retains only cross-view dependencies (no data reconstruction) and can, therefore, be markedly more data-efficient \citep{martini2024data,van2025joint}, reducing the sample demands that limit classical estimators.

We show, analytically and experimentally, that popular separable/bilinear critics \citep{oord2018representation} can systematically \emph{overestimate} task-relevant dimensionality in this pipeline, even for simple latent distributions. We mitigate this by introducing a \emph{hybrid critic} that retains an explicit $k_z$ bottleneck while allowing flexible cross-view mixing, which we find is essential for capturing nonlinear dependence geometry without inflating the embedding dimension. We also give a protocol that reads off the effective dimension from a single over-parameterized hybrid model, avoiding a sweep over $k_z$. For finite datasets we use the max-test early-stopping rule \citep{abdelaleem2025accurate} and validate the method on teacher problems with known task-relevant dimension. To enable comparison with intrinsic-dimension estimators, we extend our approach to an intrinsic setting by splitting a single dataset into two equivalent views. In noisy regimes where classical geometric estimators degrade, the hybrid MI/SIB approach remains reliable. Finally, we demonstrate utility on realistic physical data by applying the estimator to 2D Ising spins simulations (recovering critical scaling) and to single/double pendulum videos (recovering phase-space dimensionality).

\paragraph{Contributions.}
(i) Task-relevant dimensionality from paired views as a symmetric MI preservation/SIB problem; (ii) an analytic and empirical demonstration of dimensionality inflation induced by separable/bilinear critics; (iii) a hybrid critic and one-shot effective-dimension estimator with a finite-data training protocol; (iv) validation on teacher benchmarks, an intrinsic-dimension extension via view splitting, and robustness to observation noise; (v) applications to physics datasets.

\section{Setup and Methodology}

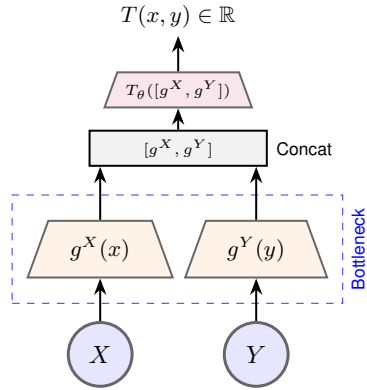
\begin{wrapfigure}{r}{0.35\textwidth}
    \centering
    \vspace{-.6in} % optional: pull it up
\centering
\resizebox{0.35\textwidth}{!}{% Resize to fit width
\begin{tikzpicture}[
    % Style definitions
    font=\small\sffamily,
    >=Stealth,
    input node/.style={circle, draw=black!60, fill=blue!10, very thick, minimum size=0.9cm, font=\bfseries},
    embedding node/.style={rectangle, draw=black!60, fill=green!10, thick, minimum width=1.1cm, minimum height=0.6cm, rounded corners},
    encoder block/.style={trapezium, draw=black!60, fill=orange!10, thick, minimum width=1.0cm, minimum height=0.8cm, trapezium angle=70},
    mlp block/.style={trapezium, draw=black!60, fill=orange!10, thick, minimum width=0.8cm, minimum height=0.8cm, trapezium angle=70},
    operation node/.style={circle, draw=black!60, fill=gray!10, thick, minimum size=0.6cm},
    label text/.style={font=\bfseries\small, align=center},
    subcap/.style={font=\bfseries, yshift=-0.5cm}
]

\begin{scope}

% Inputs
\node[input node] (x_hyb) at (-1.1, 0) {$X$};
\node[input node] (y_hyb) at (1.1, 0) {$Y$};

% Encoders
\node[encoder block, above=0.6cm of x_hyb] (enc_x_hyb) {$g^X(x)$};
\node[encoder block, above=0.6cm of y_hyb] (enc_y_hyb) {$g^Y(y)$};

\node[
    draw=blue!80, 
    dashed, 
    fit=(enc_x_hyb) (enc_y_hyb), 
    inner sep=0.35cm, 
    % Pass rotation inside the label's curly/square brackets
    label={[rotate=90, anchor=north]right:\scriptsize \textcolor{blue}{Bottleneck}}
] (bottleneck) {};

\node[above=0.3cm of enc_x_hyb, minimum width=0.7 cm, font=\scriptsize] (z_x_hyb) {$ $};
\node[above=0.3cm of enc_y_hyb, font=\scriptsize] (z_y_hyb) {$ $};

% Concat Embeddings
\node[rectangle, draw, thick, fill=gray!10, minimum width=2.5cm, minimum height=0.3cm, above=0.2cm of z_x_hyb, xshift=1.1cm, label=right:\scriptsize Concat] (concat_hyb) {\tiny $[g^X,g^Y]$};

% Head MLP
\node[mlp block, fill=purple!10, minimum width=1.0cm, minimum height=0.5,above=0.3cm of concat_hyb] (head_hyb) {\tiny $T_\theta([g^X,g^Y])$};

% Output
\node[above=0.5cm of head_hyb] (out_hyb) {$T(x,y)\in \mathbb{R}$};

% Connections
\draw[->, thick] (x_hyb) -- (enc_x_hyb);
\draw[->, thick] (y_hyb) -- (enc_y_hyb);

\draw[->, thick] (concat_hyb) -- (head_hyb);
\draw[->, thick] (head_hyb) -- (out_hyb);

\draw[->, thick] (enc_x_hyb) -- (z_x_hyb |- concat_hyb.south);
\draw[->, thick] (enc_y_hyb) -- (z_y_hyb |- concat_hyb.south);

\end{scope}

\end{tikzpicture}
}

\vspace{-8pt}  % optional: tighten space below
\caption{\textbf{Hybrid Critic Architecture.} Retains the bottleneck for dimensionality analysis, but allows flexible mixing via a concatenated head $T_\theta$ (e.g. a small MLP).}
\label{fig:critic}
    \label{fig:small}
\end{wrapfigure}

We will use mutual information (MI) between two variables $X$ and $Y$ as a measure of task relevance:
\begin{equation}
    I(X;Y)=\mathbb{E}_{p(x)}\!\left[D_{\rm KL}\!\left(p(y|x)\,\|\,p(y)\right)\right]=I(Y;X).
\end{equation}
Using the Donsker--Varadhan (DV) representation of the KL divergence~\citep{donsker1983asymptotic}, $D_{\rm KL}(P\|Q)=\sup_{T:\Omega\rightarrow\mathbb{R}}\ \mathbb{E}_P[T]-\log\!\left(\mathbb{E}_Q[e^T]\right)$, where the supremum is taken over measurable functions sharing the common support $\Omega$ of $P$ and $Q$, one obtains a variational lower bound on MI: with expectations approximated by finite-sample averages, one trades estimation of $I$ for optimization over a critic function $T(x,y)$, implemented with sufficiently expressive neural networks \citep{Belghazi2018MutualEstimation}. In this work, we largely focus on the {\em symmetrized} InfoNCE estimator \citep{oord2018representation,radford2021learning}, which is derived from DV-style bounds by {\em contrastive} averaging. It approaches the true MI when $T$ equals the optimal critic $T^*(x,y)=\log\!\big(p(x,y)/p(x)p(y)\big)$, provided the number of negatives and the batch size are large. See App.~\ref{app:theory_objectives} and \citet{abdelaleem2025accurate} for a comparison of neural MI objectives, critics, and architectures. While InfoNCE is a strong default, the exact choice of the estimator is not crucial to the rest of the story.

MI estimators also differ in how the critic is parameterized. The simplest choice is a {\em concatenated (joint)} critic, which trains a single network $T_{\rm concat}:\mathcal{X}\times\mathcal{Y}\rightarrow\mathbb{R}$ \citep{Belghazi2018MutualEstimation}. Another common choice is a {\em separable} critic \citep{oord2018representation}, which trains two encoders $g^{X}:\mathcal{X}\rightarrow\mathbb{R}^{k_z}$ and $g^{Y}:\mathcal{Y}\rightarrow\mathbb{R}^{k_z}$ and represents the critic as the bilinear form (dot product) $T_{\rm sep}(x,y)=g^{X}(x)\cdot g^{Y}(y)$. Unlike the concatenated critic, the separable form introduces an explicit $k_z$-dimensional bottleneck, making it natural to study how the estimate depends on representation size. 
With sufficiently expressive networks (and appropriate training protocols), either family can yield accurate MI estimates in high dimensions \citep{abdelaleem2025accurate}.

We note that symmetrized InfoNCE is widely used and performs well in modern representation learning \citep{chen2020simple,radford2021learning}. We therefore adopt this objective throughout. Our {\bf proposal} for estimating task-relevant dimensionality is then straightforward: use symmetrized InfoNCE with an embedding bottleneck and finite-data improvements that yield nearly unbiased estimates \citep{abdelaleem2025accurate}\footnote{See also \cite{assran2023self,Monemi2025JEPATutorial} for connections to joint-embedding architectures.}, sweep over the bottleneck dimension $k_z$, and identify the smallest $k_z$ beyond which the MI estimate no longer increases within error bars as the task-relevant dimension. Crucially, we will show that existing critic architectures fail in this pipeline: a concatenated critic has no explicit bottleneck to interpret as task-relevant dimension, while a separable critic can require \emph{inflated} $k_z$ to represent nonlinear dependencies, leading to systematic overestimation. We therefore introduce a \emph{hybrid} critic,
\begin{equation}
    T_{\rm hybrid}(x,y)=T_\theta\!\big([g^{X}(x),g^{Y}(y)]\big),
\end{equation}
which retains the $k_z$ bottleneck but uses a lightweight network $T_\theta:\mathbb{R}^{k_z}\times\mathbb{R}^{k_z}\to\mathbb{R}$ to capture nonlinear cross-view interactions without inflating $k_z$ (Figure.~\ref{fig:critic}). This decouples \emph{representation size} from \emph{critic expressivity}, enabling the learned embeddings to reflect the shared task-relevant latent dimensionality (and, as we show later, intrinsic dimensionality) without increasing $k_z$. Parenthetically, the hybrid architecture also performs well as an MI estimator in its own right.

In principle, $T_\theta$ can be a small Multilayer Perceptron (MLP) head operating in the latent space, while the encoders can be tailored to the modality (e.g., incorporating invariances or convolutional structure) to map observations into a shared latent space. Here we instead fix both the encoders and the hybrid head to the same small MLP architecture across all experiments, and show that this minimal design still recovers shared latent and intrinsic dimensionality across diverse datasets where existing estimators fail.

\section{Results}
\label{sec:results-synthetic}

We first present estimation of task-relevant dimensionality on synthetically generated datasets constructed from low-dimensional ($\mathcal{O}(1)$) latent variables mapped into a high-dimensional observation space by fixed nonlinear functions $F$ (see App.~\ref{app:details_synthetic_data}). We use the following notation: $Z_X$ and $Z_Y$ are latent variables
with joint distribution $p_Z(z_x,z_y)$ and dimensions $K_{Z_X}=\dim(Z_X)$ and
$K_{Z_Y}=\dim(Z_Y)$. Observations are $(X,Y)=(F_X(Z_X),F_Y(Z_Y))$, with
$F_X:\mathbb{R}^{K_{Z_X}}\!\to\mathbb{R}^{K_X}$ and $F_Y:\mathbb{R}^{K_{Z_Y}}\!\to\mathbb{R}^{K_Y}$.
Unless otherwise stated, we take $K_{Z_X}=K_{Z_Y}\equiv K_Z$ and $K_X=K_Y\equiv K$.

We focus on the physically relevant regime $K\gg K_Z$ and aim to infer $K_Z$ from samples of the $K$-dimensional observations $(X,Y)$. Unless otherwise stated, we fix $K=500$ and vary $K_Z$, the latent distribution $p_Z$, and the observation maps $F$, while comparing critic sizes and architectures.

Across all considered benchmarks, the \emph{hybrid} architecture reliable infers the latent dimensionality. We therefore apply the same pipeline to more realistic data, with similarly good results.

\subsection{Infinite Data Regime: high dimensional input, low dimensional latents}
\label{subsec:infinite-highdim}

We begin in the \emph{infinite-data} regime, where each optimization step uses an independently sampled batch from the data-generating distribution. This eliminates overfitting and isolates the estimator’s intrinsic behavior, rather than finite-sample effects. To use MI as a dimensionality estimator, we study the optimized MI estimate as a function of an effective representation size. For neural estimators this size is the embedding dimension $k_z$ of the encoders $g_X,g_Y:\mathbb{R}^{K}\to\mathbb{R}^{k_z}$ (we use same dimensionality encoders throughout). We also investigate MI estimation via CCA \citep{gelfand1959calculation} (also see App.~\ref{app:theory_gaussian_derivation}), where the analogous knob is the number of retained canonical pairs.

\begin{figure}[t]
\centering
\includegraphics[width=1.0\linewidth]{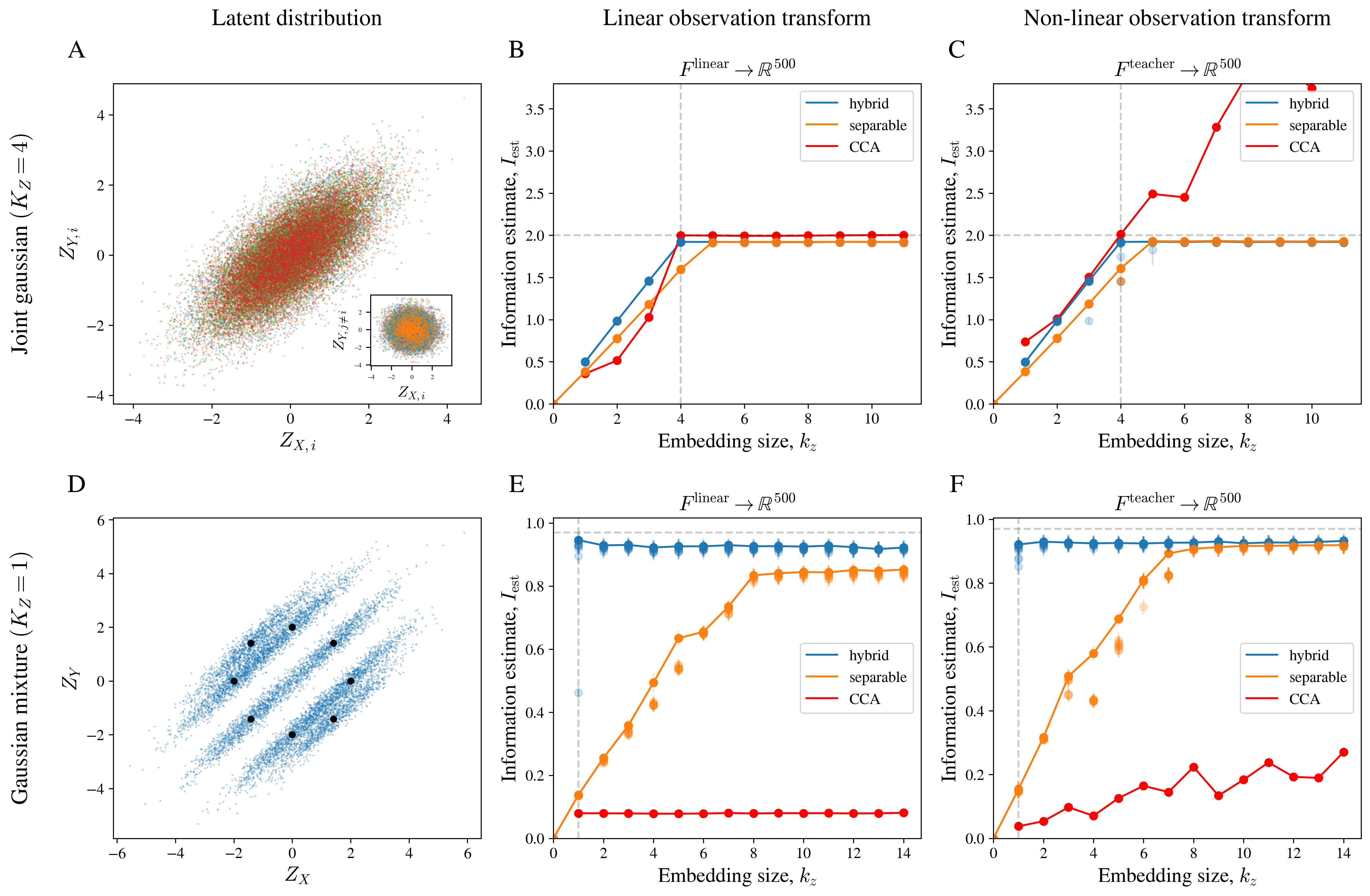}
\caption{
\textbf{Role of embedding size in the infinite-data (resampling) regime.}
Estimated MI versus encoder embedding size $k_z$ for two latent distributions:
\textbf{(A--C)} jointly Gaussian latent ($K_Z=4$) with total MI $I=2.0$ bits (equal per latent dimension);
\textbf{(D--F)} Gaussian mixture with $N_p=8$ equally likely joint-Gaussian clusters with $k_Z=1$
(each with $\rho\approx 0.97$), with cluster means on a circle of radius $\mu=2.0$ (see App.~\ref{app:details_synthetic_data}).
\textbf{(A,D)} Latent distributions. \textbf{(B,E)} MI estimates (maximum over 10 trials, individual trials shown with semi-transparent markers) for frozen
linear observation maps. \textbf{(C,F)} Same, but with frozen nonlinear teacher maps (see
App.~\ref{app:details_synthetic_data}). Vertical dotted lines mark true task-relevant dimension, which is always matched by $k_z^\ast$ chosen by the hybrid critic.}
\label{fig:fig2_infinite_synthetic_joint}
\end{figure}

For sufficiently large $k_z$, the MI estimate generically saturates to the true MI of the latent distribution. We denote the saturation location by $k_z^\ast$. In general, $k_z^\ast$ depends on the latent distribution $p_Z$, the true latent dimensionality $K_Z$, and the estimator/critic architecture. Figure~\ref{fig:fig2_infinite_synthetic_joint} illustrates this behavior for two representative latent distributions (many more were tested) used throughout this section: (i) a jointly Gaussian latent with total MI $I=2.0$ bits and true shared latent dimensionality $K_Z=4$, Fig.~\ref{fig:fig2_infinite_synthetic_joint}A, and (ii) a deliberately challenging, multimodal Gaussian mixture with $K_Z=1$ and multiple equally likely clusters, Fig.~\ref{fig:fig2_infinite_synthetic_joint}D. In each case we generate observations using either frozen linear maps or frozen nonlinear teacher networks.

In the infinite-data regime, CCA succeeds only for the jointly Gaussian latent with linear
observations (Fig.~\ref{fig:fig2_infinite_synthetic_joint}B). With nonlinear observation maps, CCA
picks up spurious correlations and fails to recover both MI and latent dimensionality in both benchmarks  (Fig.~\ref{fig:fig2_infinite_synthetic_joint}C,E,F).
The \emph{separable} critic can estimate MI but does not saturate at the correct dimensionality: for
the joint Gaussian, saturation occurs at $k_z^\ast=K_Z+1$
(Fig.~\ref{fig:fig2_infinite_synthetic_joint}B,C;  App.~\ref{app:theory_solvable}), while the
mixture yields substantial overestimation (Fig.~\ref{fig:fig2_infinite_synthetic_joint}E,F; 
App.~\ref{app:non_invert_mix}). In contrast, for the \emph{hybrid} critic we find $k_z^\ast=K_Z$
across both latent distributions and both observation transforms, making $k_z^\ast$ a reliable
estimate of the shared task-relevant latent dimensionality. We therefore focus on the hybrid critic
in the remainder of this work.

\subsection{Noise in the observation space and intrinsic dimensionality}
\label{subsec:infinite-noise}

We target our estimator to scientific applications, where data are always noisy. We thus
test robustness to measurement noise by adding  observation noise after the (frozen)
observation maps:
\begin{equation}
    X = F_X(Z_X) + \eta_X,\qquad
    Y = F_Y(Z_Y) + \eta_Y,
\end{equation}
with $\eta_X,\eta_Y$ uncorrelated Gaussian white noise,
$\langle \eta_{\alpha,i}\eta_{\beta,j}\rangle=\sigma_\alpha^2\delta_{ij}\delta_{\alpha\beta}$ for
$\alpha,\beta\in\{X,Y\}$ and $i,j\in\{1,\ldots,K\}$. Noise reduces the achievable MI between $X$ and
$Y$, and we test whether the saturation point $k^\ast_z$ remains controlled by the shared latent
dimensionality rather than by noise.

\begin{figure}[t]
    \centering
\includegraphics[width=0.7\linewidth]{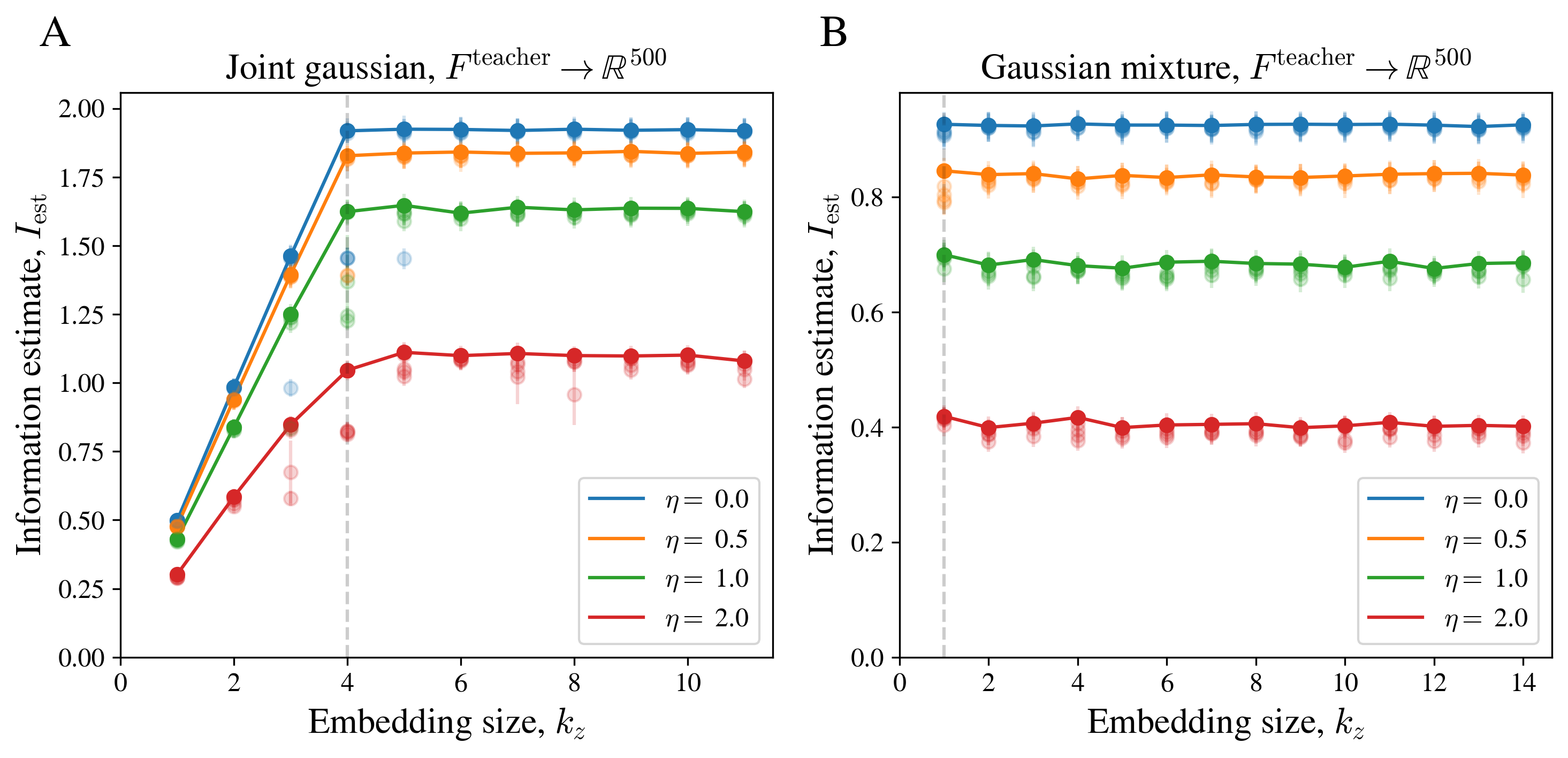}
\caption{
\textbf{Effect of additive observation noise (hybrid critic).}
Independent white noise is added after the frozen nonlinear observation map, with
$\langle \eta_{\alpha,i}\eta_{\beta,j}\rangle=\sigma_\alpha^2\delta_{ij}\delta_{\alpha\beta}$ and
strength set by the noise-to-signal ratio $\eta$.
\textbf{(A)} Joint Gaussian latent ($K_Z=4$).
\textbf{(B)} Gaussian mixture ($N_p=8$ components).
Noise reduces the estimated MI, while the saturation point used to infer dimensionality is
preserved. MI is the maximum over 10 trials; individual trials shown as semi-transparent
markers.
}
\label{fig:fig3_infinte_synthetic_joint_noisy}
\end{figure}

Figure~\ref{fig:fig3_infinte_synthetic_joint_noisy} shows the optimized MI estimate versus embedding
size for the same two representative latent distributions as in
Fig.~\ref{fig:fig2_infinite_synthetic_joint}. We parameterize the noise level by the noise-to-signal
ratio $\eta := \sqrt{{\sigma_X^2}/{\mathrm{var}(F_X(Z_X))}} = \sqrt{{\sigma_Y^2}/{\mathrm{var}(F_Y(Z_Y))}}$.
Increasing noise lowers the estimated MI, but the dimensionality inferred from the saturation point
$k_z^\ast$ is unchanged across the range shown for both latent distributions.

The same behavior holds in a shared-latent setting, where two noisy views are generated from a
single latent variable (Appendix~\ref{app:shared_latent}, Fig.~\ref{fig:shared_latent}), motivating
our later ``view-splitting'' constructions for intrinsic-dimension benchmarks and real data. For
comparison, standard intrinsic-dimension estimators such as Levina--Bickel and Two-NN
\citep{levina2004maximum,facco2017estimating} degrade sharply under observation noise and tend to
track the ambient observation dimension rather than the latent one (Appendix~\ref{app:compare_with_ID},
Fig.~\ref{fig:compare_with_id}); in the same regime, our hybrid critic MI-based estimator continues to
recover the shared latent dimensionality.

\subsection{Single-shot dimensionality estimation via participation ratio}
\label{subsec:infinite-pr-dim}

So far we inferred latent dimensionality from the saturation of the optimized MI estimate as a
function of bottleneck size $k_z$ for the hybrid critic. Sweeping $k_z$ is computationally
expensive and often impractical. Empirically, the variability of the learned MI estimates across
runs decreases as $k_z$ increases beyond $K_Z$ (cf.~Figs.~\ref{fig:fig2_infinite_synthetic_joint}E,F
and~\ref{fig:fig3_infinte_synthetic_joint_noisy}A), suggesting more stable optimization under
overparameterization. This raises a natural question: when $k_z>K_Z$, do the learned embeddings
actually use the additional dimensions, or does the cross-view signal concentrate onto an
effectively $K_Z$-dimensional subspace?

We quantify this by the cross-covariance of the encoder outputs,
$$ C_{xy} = \left(g_X(X) - \bar g_X(X)\right)^T \left(g_Y(Y) - \bar g_Y(Y)\right), \qquad
\bar g_X = \langle g_X \rangle,\;\;
\bar g_Y = \langle g_Y \rangle,$$
and define an effective dimension from its singular values $\{\sigma_i\}_{i=1}^{k_z}$ via the
participation ratio
\begin{equation}
d_{\mathrm{eff}}=\frac{\left(\sum_i \sigma_i\right)^2}{\sum_i \sigma_i^2}.
\label{eq:deff_defintion}
\end{equation}
This choice soft-counts dominant modes while suppressing the noise tail (see
App.~\ref{app:protocol_pr}).

\begin{figure}[t]
    \centering
    \includegraphics[width=1.0\linewidth]{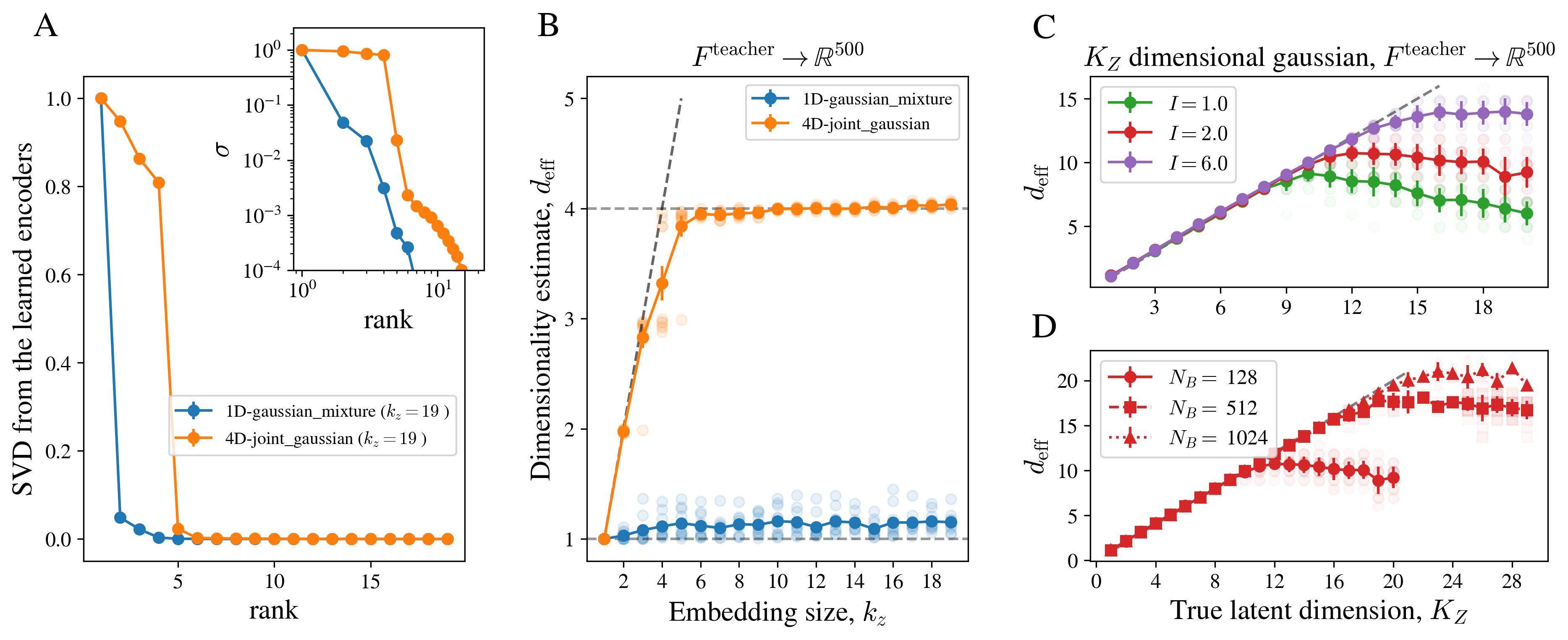}
    \caption{\textbf{Single-shot dimensionality from the embedding spectrum (hybrid critic).}
The participation ratio of the cross-covariance spectrum of the learned encoder embeddings provides a
reliable estimate of latent dimensionality.
\textbf{(A)} Normalized singular values of the cross-covariance (computed from $10^4$ samples) for a
trained model with $k_z=19$ (inset: log-scale). A clear gap appears after the first $K_Z$ modes, yielding
$d_{\rm eff}$ via Eq.~\ref{eq:deff_defintion}.
\textbf{(B)} $d_{\rm eff}$ saturates for $k_z\gtrsim K_Z$ for both representative latent distributions,
indicating that the learned embeddings concentrate onto an effectively $K_Z$-dimensional subspace.
\textbf{(C,D)} $d_{\rm eff}$ from a single over-parameterized model ($k_z=64$) versus $K_Z$ for jointly
Gaussian latents: varying total MI at fixed batch size $N_B=128$ (C), and varying $N_B$ at fixed total
MI $I=2$ bits (D). As always, semi-transparent markers denote individual trials, and error bars are standard deviations. }
    \label{fig:fig5_pr_infinite_synthetic}
\end{figure}

Figure~\ref{fig:fig5_pr_infinite_synthetic}A shows that for $k_z\gg K_Z$  the embedding spectrum exhibits a
clear gap beyond rank $K_Z$ for both representative latent distributions, yielding $d_{\rm eff}$
from a single trained model. As shown in Fig.~\ref{fig:fig5_pr_infinite_synthetic}B, $d_{\rm eff}$
is stable for $k_z\gtrsim K_Z$, with variance of $d_{\rm{eff}}$ across trials decreasing as $k_z$ increases, again reflecting more stable training.

This yields a dimensionality estimator without a $k_z$ sweep: choose $k_z$ sufficiently large, train
one hybrid MI estimator, and report $d_{\rm eff}$ from the embedding spectrum (in practice, take
$k_z-d_{\rm eff}\gg 1$). Figures~\ref{fig:fig5_pr_infinite_synthetic}C,D show exact recovery
$d_{\rm eff}=K_Z$ for jointly Gaussian latents with varying $K_Z$ observed through a nonlinear map into $K=500$
dimensions using a fixed $k_z=64$. Deviations arise only when finite-batch effects dominate the
latent signal, either because the signal per latent dimension is small
(Fig.~\ref{fig:fig5_pr_infinite_synthetic}C) or because the batch size $N_B$ is insufficient to
sample all latent directions (Fig.~\ref{fig:fig5_pr_infinite_synthetic}D).

\subsection{Finite data}
\label{subsec:finite-pr-dim}

\begin{figure}[t]
    \centering
    \includegraphics[width=1.0\linewidth]{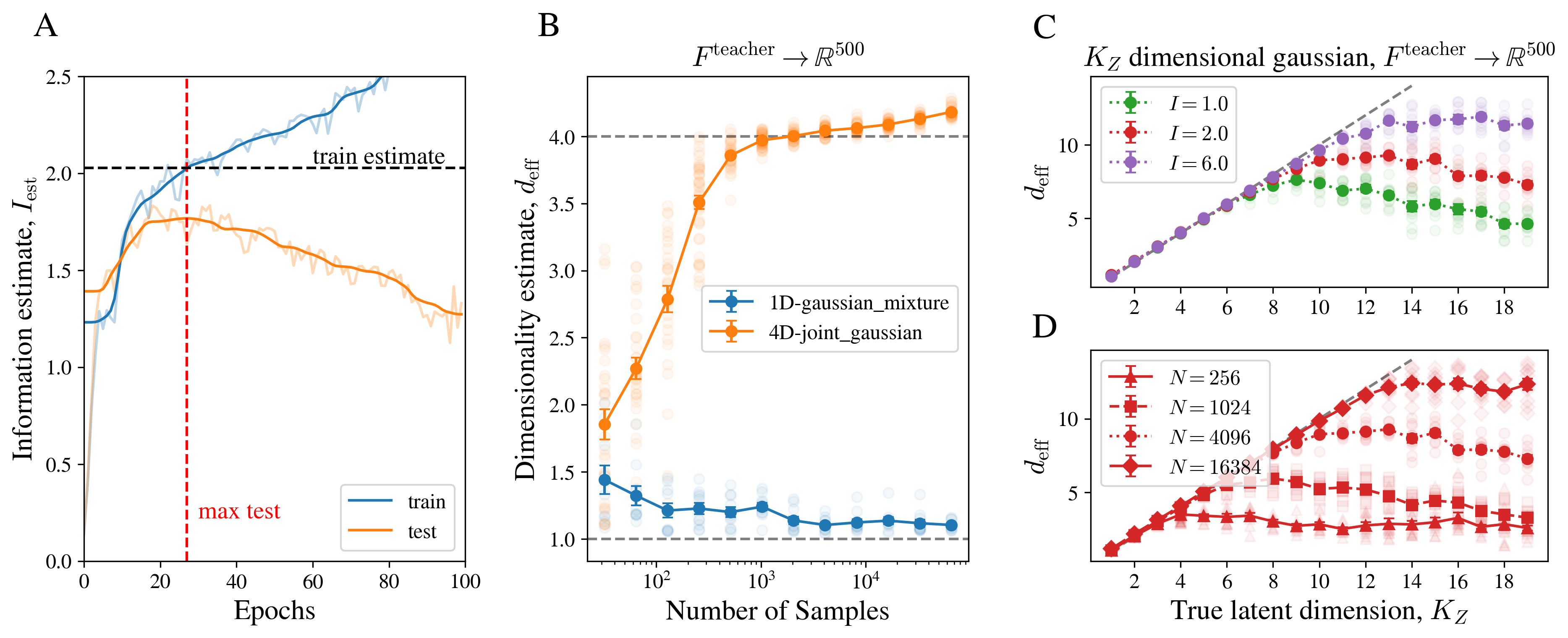}
    \caption{\textbf{Dimensionality estimation with finite data.}
\textbf{(A)} Max-test, train-estimate protocol \citep{abdelaleem2025accurate} for a jointly Gaussian
latent ($K_Z=4$, $I=2$ bits) observed through a teacher map into $\mathbb{R}^{500}$ with $N=1024$
samples: select $t^\ast=\arg\max_t \widehat I_{\mathrm{test}}(t)$ and report $\widehat I_{\mathrm{train}}(t^\ast)$.
Solid MI curves shown (and used to locate the maximum) are median-filtered over 20 epochs.
\textbf{(B)} $d_{\rm eff}$ from the trained encoders ($k_z=64$) versus number of training samples for
the two representative latent distributions (cross-covariance computed on the full training set).
\textbf{(C,D)} $d_{\rm eff}$ from a single model with $k_z=64$ versus $K_Z$ for jointly Gaussian latents:
varying total MI at fixed $N=4096$ (C), and varying $N$ at fixed $I=2$ bits (D). Usual notation for markers/error bars is used.}
    \label{fig:fig6_finite_synthetic}
\end{figure}

With a fixed finite dataset, variational MI bounds can overfit, so dimensionality estimation requires
an explicit early-stopping rule. We use the \emph{max-test, train-estimate} protocol of
\citet{abdelaleem2025accurate} (Fig.~\ref{fig:fig6_finite_synthetic}A): we select the training checkpoint
$t^\ast=\arg\max_t \widehat I_{\mathrm{test}}(t)$, but report $\widehat I_{\mathrm{train}}(t^\ast)$ as the MI estimate.
This choice is appropriate because our objective is estimating $I(X;Y)$ from the available sample,
not optimizing predictive generalization, and it empirically reduces bias from overfitting. The same checkpoint yields encoder representations used for the
participation-ratio estimate $d_{\rm eff}$.

Figure~\ref{fig:fig6_finite_synthetic}B shows that $d_{\rm eff}$ converges to the correct value for
both representative latent distributions with $\sim10^3$ samples. Figures~\ref{fig:fig6_finite_synthetic}C,D
show $d_{\rm eff}$ versus the true latent dimension $K_Z$ for jointly Gaussian latents across MI
levels and dataset sizes. Our task-relevant dimensionality estimator remains accurate over a broad range
of $K_Z$, failing only when the sample size is insufficient to resolve the latent signal. Taken together,
these results show that latent dimensionality can be inferred from finite datasets using sufficiently
expressive encoders, provided overfitting of MI estimates is controlled.

Our complete task-relevant dimensionality estimation protocol for finite data is:

\begin{tcolorbox}[breakable,boxrule=0.8pt,arc=2pt,left=6pt,right=6pt,top=4pt,bottom=4pt]
\textbf{Task-relevant dimensionality estimation (finite data).} Given paired observations $(X,Y)$:
\begin{enumerate}[leftmargin=15pt,itemsep=0.2em,topsep=0.2em]
  \item Choose $k_z$ larger than the expected task-relevant dimension.
  \item Train an MI estimator with the hybrid critic using the max-test, train-estimate heuristic.
  \item Compute the cross-covariance of the learned encoder representations.
  \item Estimate dimensionality via the participation ratio $d_{\rm eff}$ of the singular-value spectrum.
\end{enumerate}
\end{tcolorbox}

\subsection{Estimating task-relevant dimensionality of physics datasets}
\label{subsec:results-real}
We now apply our dimensionality estimation protocol to more realistic datasets (simulated and
experimental) whose latent structure is constrained by well-understood physics. In each case, we
use the same estimator, training protocol, and network architectures as in the synthetic benchmarks,
 and test whether the inferred dimensionality reflects meaningful physical
structure.

\subsubsection{Ising model}
Machine learning has been used to detect critical behavior in
statistical mechanics with varying degrees of success
\citep{carrasquilla2017machine,giannetti2019machine,carleo2019machine}. Here we show that our
dimensionality estimator identifies the phase transition and recovers the expected scaling
in the 2D Ising model. In this system, near criticality, correlated domains grow and the correlation length diverges as
$\xi\sim|T-T_c|^{-\nu}$, with $T_c=2/\ln(1+\sqrt{2})\approx 2.269$ and $\nu=1$
\citep{goldenfeld2018lectures}. The typical number of effectively independent domains spanning the shared interface scales as $\sim (L/\xi)$, suggesting that $d_{\rm eff}$ should follow a finite-size scaling form and collapse
across system sizes when plotted against the scaling variable $L/\xi\sim L/|T/T_c-1|^{-\nu}$, with
$\nu=1$.

We simulate the Ising model on a periodic $L\times L$ lattice with $L\in[13,133]$ (see
App.~\ref{app:details_ising}). To construct paired views from a single configuration, we use the
view-splitting procedure of App.~\ref{app:shared_latent}, taking the upper-left half of spins as $X$
and the lower-right half as $Y$, and apply the same estimator and training protocol as in the
synthetic benchmarks. The recovered effective dimensionality $d_{\rm eff}$ tracks the growth of
correlated domains (Fig.~\ref{fig:ising_supplementary}) and exhibits a clear scaling collapse across
system sizes (Fig.~\ref{fig:real_data}A). The inferred
dimensionality is therefore not an artifact of architecture or tuning, but reflects physically
meaningful collective structure in the underlying system.

\subsubsection{Pendulum dynamics}
We next consider video recordings of simple mechanical systems and ask whether our estimator can
recover the number of underlying degrees of freedom directly from raw pixels. Here $X$ is the delayed embedding \citep{takens2006detecting} of two sequential ``past'' video frames, and $Y$ is the same of two future frames, so that the task-relevant dimensionality measures the number of degrees of freedom needed to predict future from the past. We analyze movies of
a single pendulum (two degrees of freedom) and a chaotic double pendulum (four degrees of freedom),
from \citet{chen2022automated} (see App.~\ref{app:details_pendulum}). We then apply our standard pipeline to infer the dimensionality of the underlying 
dynamics.

\begin{figure}[t]
    \centering
    \includegraphics[width=0.9\linewidth]{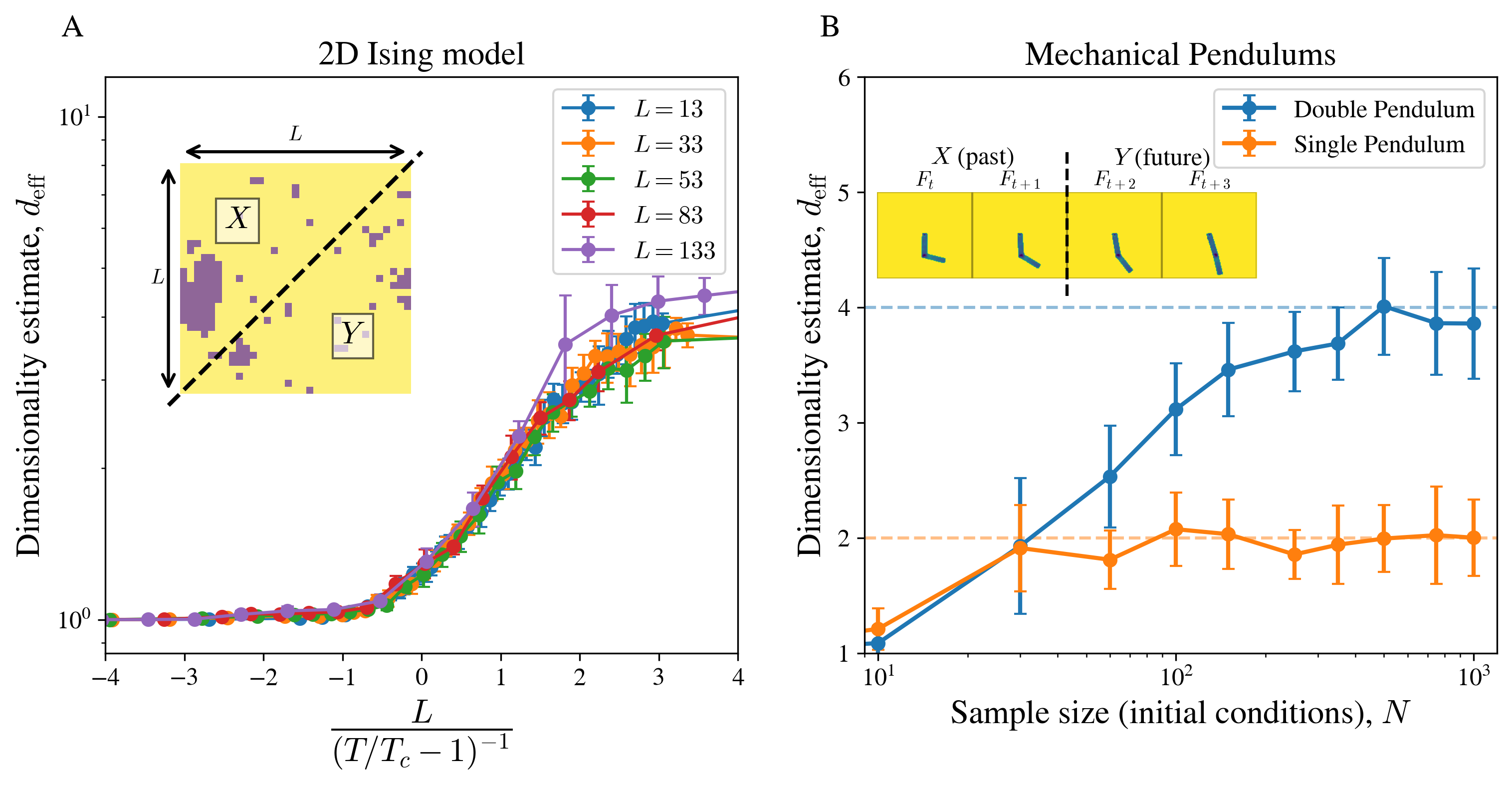}
    \caption{\textbf{Dimensionality estimation on physical datasets.}
    We apply the same protocol as in the synthetic benchmarks to (A) 2D nearest-neighbor Ising
    configurations ($J=1$) and (B) single/double pendulum videos from \citet{chen2022automated}; error bars are standard deviations over $10$ trials each.
    \textbf{(A)} $10^4$ MCMC samples on an $L\times L$ lattice; paired views $(X,Y)$ are formed by a
    spatial split of each configuration (inset). The measured $d_{\rm eff}$ exhibits finite-size
    scaling and collapses across $L$ when plotted against the standard scaling variable
    $L/(T/T_c-1)^{-1}$ with $\nu=1$, with the collapse centered at $T_c$.
    \textbf{(B)} Paired views are constructed from the temporal structure of each video by stacking
    past and future frames (inset). With sufficient samples, $d_{\rm eff}$ recovers the expected
    phase-space dimensions for the single (2) and double (4) pendulum.}
    \label{fig:real_data}
\end{figure}

For this system, autoencoder based approached for dimensionality estimation have proven to be brittle \citep{chen2022automated}. In contrast, our approach reliably infers the expected phase-space dimensions for both systems (Fig.~\ref{fig:real_data}B) from as few as a hundred samples, even in the chaotic pendulum.

\section{Discussion}
We proposed a method to measure \emph{task-relevant} dimensionality: the minimal latent-state
dimension needed to preserve the information required for a specified prediction problem. In our
setting, relevance is defined by shared structure between paired views, quantified by MI. Thus
``dimension'' depends on the task and on how views are constructed (past--future or spatial splits),
and is not a vague, scale-dependent intrinsic property of the raw observations.

Standard neural MI estimators are not automatically usable as dimensionality estimators. For
example, separable/bilinear critics can systematically overestimate the dimension even in simple
cases. Our {\em hybrid} critic addresses this by decoupling the interpretable bottleneck size (data
geometry) from critic expressivity (architectural constraints). We also introduced a one-shot
estimator based on the participation ratio of the cross-covariance spectrum in embedding space,
combined with an early-stopping protocol that controls overfitting of variational MI bounds. This
avoids sweeping $k_z$ and yields a stable estimate once the model is sufficiently overparameterized.

Limitations remain: the estimate depends on the success of the underlying MI estimation, on view construction, on
the choice of embedding-network architecture, and on having enough samples to resolve latent
structure. It will therefore be important to formalize conditions under which it succeeds or fails
beyond the settings explored here. Nonetheless, unlike neighbor-scaling intrinsic-dimension
estimators, our approach does not degrade quickly under observation noise, making it particularly
useful for analyzing the latent representation geometry of noisy experimental datasets. The physics case studies support
this: the resulting $d_{\rm eff}$ exhibits finite-size scaling near the 2D Ising critical point and
recovers phase-space dimensions from raw pendulum videos.

% \newpage

\subsubsection*{Acknowledgments}
We thank Sean Ridout for providing MCMC code to generate Ising spin configurations and for critical discussions. We thank K.\ Michael Martini for many discussions over the years. PG was funded, in part, by the Tarbutton Interdisciplinary Postdoctoral Fellowship at Emory College of Arts and Sciences. PG and IN were funded, in part, by the Simons Fondation Investigator grant to IN. EA and AS were supported, in part, by NIH Grant No. RF1-MH130413 and Brain and Behavior Research Foundation Young Investigator Grant 30885. We acknowledge support of our work through the use of the HyPER C3 cluster of Emory University’s AI.Humanity Initiative.

\bibliography{ref}

@article{abdelaleem2025accurate,
  title={Accurate Estimation of Mutual Information in High Dimensional Data},
  author={Abdelaleem, Eslam and Martini, K Michael and Nemenman, Ilya},
  journal={arXiv preprint arXiv:2506.00330},
  year={2025}
}

@article{mardt2018vampnets,
  title={VAMPnets for deep learning of molecular kinetics},
  author={Mardt, Andreas and Pasquali, Luca and Wu, Hao and No{\'e}, Frank},
  journal={Nature communications},
  volume={9},
  number={1},
  pages={5},
  year={2018},
  publisher={Nature Publishing Group UK London}
}

@article{meng2022compressed,
  title={Compressed predictive information coding},
  author={Meng, Rui and Luo, Tianyi and Bouchard, Kristofer},
  journal={arXiv preprint arXiv:2203.02051},
  year={2022}
}

@article{schmid2022dynamic,
  title={Dynamic mode decomposition and its variants},
  author={Schmid, Peter J},
  journal={Annual Review of Fluid Mechanics},
  volume={54},
  number={1},
  pages={225--254},
  year={2022},
  publisher={Annual Reviews}
}

@article{lecun2022path,
  title={A path towards autonomous machine intelligence version 0.9. 2, 2022-06-27},
  author={LeCun, Yann},
  journal={Open Review},
  volume={62},
  number={1},
  pages={1--62},
  year={2022}
}

@article{giannetti2019machine,
  title={Machine learning as a universal tool for quantitative investigations of phase transitions},
  author={Giannetti, Cinzia and Lucini, Biagio and Vadacchino, Davide},
  journal={Nuclear Physics B},
  volume={944},
  pages={114639},
  year={2019},
  publisher={Elsevier}
}

@inproceedings{takens2006detecting,
  title={Detecting strange attractors in turbulence},
  author={Takens, Floris},
  booktitle={Dynamical Systems and Turbulence, Warwick 1980: proceedings of a symposium held at the University of Warwick 1979/80},
  pages={366--381},
  year={2006},
  organization={Springer}
}

@book{goldenfeld2018lectures,
  title={Lectures on phase transitions and the renormalization group},
  author={Goldenfeld, Nigel},
  year={2018},
  publisher={CRC Press}
}

@article{carrasquilla2017machine,
  title={Machine learning phases of matter},
  author={Carrasquilla, Juan and Melko, Roger G},
  journal={Nature Physics},
  volume={13},
  number={5},
  pages={431--434},
  year={2017},
  publisher={Nature Publishing Group UK London}
}

@article{carleo2019machine,
  title={Machine learning and the physical sciences},
  author={Carleo, Giuseppe and Cirac, Ignacio and Cranmer, Kyle and Daudet, Laurent and Schuld, Maria and Tishby, Naftali and Vogt-Maranto, Leslie and Zdeborov{\'a}, Lenka},
  journal={Reviews of Modern Physics},
  volume={91},
  number={4},
  pages={045002},
  year={2019},
  publisher={APS}
}

@article{gelfand1959calculation,
  title={Calculation of the amount of information about a random function contained in another such function},
  author={Gelfand, Izrail Moiseevich},
  journal={Amer. Math. Soc. Transl.},
  volume={12},
  pages={199--246},
  year={1959}
}

@article{bialek2001predictability,
  title={Predictability, complexity, and learning},
  author={Bialek, William and Nemenman, Ilya and Tishby, Naftali},
  journal={Neural computation},
  volume={13},
  number={11},
  pages={2409--2463},
  year={2001},
  publisher={MIT Press}
}

@article{tchernookov2013predictive,
  title={Predictive information in a nonequilibrium critical model},
  author={Tchernookov, Martin and Nemenman, Ilya},
  journal={Journal of Statistical Physics},
  volume={153},
  number={3},
  pages={442--459},
  year={2013},
  publisher={Springer}
}

@inproceedings{assran2023self,
  title={Self-supervised learning from images with a joint-embedding predictive architecture},
  author={Assran, Mahmoud and Duval, Quentin and Misra, Ishan and Bojanowski, Piotr and Vincent, Pascal and Rabbat, Michael and LeCun, Yann and Ballas, Nicolas},
  booktitle={Proceedings of the IEEE/CVF Conference on Computer Vision and Pattern Recognition},
  pages={15619--15629},
  year={2023}
}

@misc{Monemi2025JEPATutorial,
  author       = {Monemi, Mehdi and Chinipardaz, Maryam and Rasti, Mehdi and Bennis, Mehdi and Latva-aho, Matti},
  title        = {Tutorial on Joint Embedding Predictive Architectures ({JEPA}): {F}oundations, Applications, and Future Directions},
  year         = {2025},
  month        = dec,
  note         = {TechRxiv preprint, v1},
  doi          = {10.36227/techrxiv.176469421.19270944/v1},
}

@article{gallego2017neural,
  title={Neural manifolds for the control of movement},
  author={Gallego, Juan A and Perich, Matthew G and Miller, Lee E and Solla, Sara A},
  journal={Neuron},
  volume={94},
  number={5},
  pages={978--984},
  year={2017},
  publisher={Elsevier}
}

@article{theiler1986spurious,
  title={Spurious dimension from correlation algorithms applied to limited time-series data},
  author={Theiler, James},
  journal={Physical review A},
  volume={34},
  number={3},
  pages={2427},
  year={1986},
  publisher={APS}
}

@article{martini2024data,
  title={Data efficiency, dimensionality reduction, and the generalized symmetric information bottleneck},
  author={Martini, K Michael and Nemenman, Ilya},
  journal={Neural Computation},
  volume={36},
  number={7},
  pages={1353--1379},
  year={2024},
  publisher={MIT Press 255 Main Street, 9th Floor, Cambridge, Massachusetts 02142, USA~…}
}

@article{friedman2013multivariate,
  title={Multivariate information bottleneck},
  author={Friedman, Nir and Mosenzon, Ori and Slonim, Noam and Tishby, Naftali},
  journal={arXiv preprint arXiv:1301.2270},
  year={2013}
}

@article{van2025joint,
  title={Joint Embedding vs Reconstruction: Provable Benefits of Latent Space Prediction for Self Supervised Learning},
  author={Van Assel, Hugues and Ibrahim, Mark and Biancalani, Tommaso and Regev, Aviv and Balestriero, Randall},
  journal={arXiv preprint arXiv:2505.12477},
  year={2025}
}

@article{huh2024platonic,
  title={The platonic representation hypothesis},
  author={Huh, Minyoung and Cheung, Brian and Wang, Tongzhou and Isola, Phillip},
  journal={arXiv preprint arXiv:2405.07987},
  year={2024}
}

@article{eckmann1992fundamental,
  title={Fundamental limitations for estimating dimensions and Lyapunov exponents in dynamical systems},
  author={Eckmann, J-P and Ruelle, David},
  journal={Physica D: Nonlinear Phenomena},
  volume={56},
  number={2-3},
  pages={185--187},
  year={1992},
  publisher={Elsevier}
}

@article{jin2021jamming,
  title={A jamming plane of sphere packings},
  author={Jin, Yuliang and Yoshino, Hajime},
  journal={Proceedings of the National Academy of Sciences},
  volume={118},
  number={14},
  pages={e2021794118},
  year={2021},
  publisher={National Academy of Sciences}
}

@article{cubuk2017structure,
  title={Structure-property relationships from universal signatures of plasticity in disordered solids},
  author={Cubuk, Ekin Dogus and Ivancic, RJS and Schoenholz, Samuel S and Strickland, DJ and Basu, Anindita and Davidson, ZS and Fontaine, Julien and Hor, Jyo Lyn and Huang, Y-R and Jiang, Y and others},
  journal={Science},
  volume={358},
  number={6366},
  pages={1033--1037},
  year={2017},
  publisher={American Association for the Advancement of Science}
}

@article{tamura2022structural,
  title={Structural analysis based on unsupervised learning: Search for a characteristic low-dimensional space by local structures in atomistic simulations},
  author={Tamura, Ryo and Matsuda, Momo and Lin, Jianbo and Futamura, Yasunori and Sakurai, Tetsuya and Miyazaki, Tsuyoshi},
  journal={Physical Review B},
  volume={105},
  number={7},
  pages={075107},
  year={2022},
  publisher={APS}
}

@article{Wigner1960,
  author  = {Wigner, Eugene P.},
  title   = {The Unreasonable Effectiveness of Mathematics in the Natural Sciences},
  journal = {Communications on Pure and Applied Mathematics},
  volume  = {13},
  number  = {1},
  pages   = {1--14},
  year    = {1960},
  note    = {Richard Courant lecture in mathematical sciences delivered at New York University, May 11, 1959}
}

@article{semedo2019cortical,
  title={Cortical areas interact through a communication subspace},
  author={Semedo, Jo{\~a}o D and Zandvakili, Amin and Machens, Christian K and Yu, Byron M and Kohn, Adam},
  journal={Neuron},
  volume={102},
  number={1},
  pages={249--259},
  year={2019},
  publisher={Elsevier}
}

@article{chen2022automated,
  title={Automated discovery of fundamental variables hidden in experimental data},
  author={Chen, Boyuan and Huang, Kuang and Raghupathi, Sunand and Chandratreya, Ishaan and Du, Qiang and Lipson, Hod},
  journal={Nature Computational Science},
  volume={2},
  number={7},
  pages={433--442},
  year={2022},
  publisher={Nature Publishing Group US New York}
}

@article{das2006low,
  title={Low-dimensional, free-energy landscapes of protein-folding reactions by nonlinear dimensionality reduction},
  author={Das, Payel and Moll, Mark and Stamati, Hernan and Kavraki, Lydia E and Clementi, Cecilia},
  journal={Proceedings of the National Academy of Sciences},
  volume={103},
  number={26},
  pages={9885--9890},
  year={2006},
  publisher={National Academy of Sciences}
}

@article{schneider2023learnable,
  title={Learnable latent embeddings for joint behavioural and neural analysis},
  author={Schneider, Steffen and Lee, Jin Hwa and Mathis, Mackenzie Weygandt},
  journal={Nature},
  volume={617},
  number={7960},
  pages={360--368},
  year={2023},
  publisher={Nature Publishing Group UK London}
}

@article{camastra2016intrinsic,
  title={Intrinsic dimension estimation: Advances and open problems},
  author={Camastra, Francesco and Staiano, Antonino},
  journal={Information Sciences},
  volume={328},
  pages={26--41},
  year={2016},
  publisher={Elsevier}
}

@article{grassberger1983measuring,
  title={Measuring the strangeness of strange attractors},
  author={Grassberger, Peter and Procaccia, Itamar},
  journal={Physica D: nonlinear phenomena},
  volume={9},
  number={1-2},
  pages={189--208},
  year={1983},
  publisher={Elsevier}
}

@article{levina2004maximum,
  title={Maximum likelihood estimation of intrinsic dimension},
  author={Levina, Elizaveta and Bickel, Peter},
  journal={Advances in neural information processing systems},
  volume={17},
  year={2004}
}

@inproceedings{chen2020simple,
  title={A simple framework for contrastive learning of visual representations},
  author={Chen, Ting and Kornblith, Simon and Norouzi, Mohammad and Hinton, Geoffrey},
  booktitle={International conference on machine learning},
  pages={1597--1607},
  year={2020},
  organization={PmLR}
}

@article{abdelaleem2025deep,
  title={Deep Variational Multivariate Information Bottleneck-A Framework for Variational Losses},
  author={Abdelaleem, Eslam and Nemenman, Ilya and Martini, K Michael},
  journal={Journal of Machine Learning Research},
  volume={26},
  number={140},
  pages={1--50},
  year={2025}
}

@article{
abdelaleem2024simultaneous,
title={Simultaneous Dimensionality Reduction: A Data Efficient Approach for Multimodal Representations Learning},
author={Eslam Abdelaleem and Ahmed Roman and K. Michael Martini and Ilya Nemenman},
journal={Transactions on Machine Learning Research},
issn={2835-8856},
year={2024},
url={},
note={}
}

@inproceedings{Belghazi2018MutualEstimation,
    title = {{Mutual information neural estimation}},
    year = {2018},
    booktitle = {35th International Conference on Machine Learning, ICML 2018},
    author = {Belghazi, Mohamed Ishmael and Baratin, Aristide and Rajeswar, Sai and Ozair, Sherjil and Bengio, Yoshua and Courville, Aaron and Hjelm, R Devon},
    pages = {864--873},
    volume = {2},
    isbn = {9781510867963},
    arxivId = {arXiv:1801.04062v4}
}

@article{song2019understanding,
  title={Understanding the limitations of variational mutual information estimators},
  author={Song, Jiaming and Ermon, Stefano},
  journal={arXiv preprint arXiv:1910.06222},
  year={2019}
}

@inproceedings{poole2019variational,
  title={On variational bounds of mutual information},
  author={Poole, Ben and Ozair, Sherjil and Van Den Oord, Aaron and Alemi, Alex and Tucker, George},
  booktitle={Int.\ Conf.\ Machine Learning},
  pages={5171--5180},
  year={2019},
  organization={PMLR}
}

@article{nguyen2010estimating,
  title={Estimating divergence functionals and the likelihood ratio by convex risk minimization},
  author={Nguyen, XuanLong and Wainwright, Martin J and Jordan, Michael I},
  journal={IEEE Trans.\ Information Theory},
  volume={56},
  number={11},
  pages={5847--5861},
  year={2010},
  publisher={IEEE}
}

@article{oord2018representation,
  title={Representation learning with contrastive predictive coding},
  author={van den Oord, Aaron and Li, Yazhe and Vinyals, Oriol},
  journal={arXiv preprint arXiv:1807.03748},
  year={2018}
}

@article{gowri2024approximating,
  title={Approximating mutual information of high-dimensional variables using learned representations},
  author={Gowri, Gokul and Lun, Xiao-Kang and Klein, Allon M and Yin, Peng},
  journal={Advances in Neural Information Processing Systems},
  volume={37},
  pages={132843--132875},
  year={2024}
}

@article{donsker1983asymptotic,
  title={Asymptotic evaluation of certain Markov process expectations for large time. IV},
  author={Donsker, Monroe D and Varadhan, SR Srinivasa},
  journal={Communications on pure and applied mathematics},
  volume={36},
  number={2},
  pages={183--212},
  year={1983},
  publisher={Wiley Online Library}
}

@book{kullback1959information,
  title={Information theory and statistics},
  author={Kullback, S},
  year={1959},
  publisher = {John Wiley \& Sons},
  address={New York, NY}
}

@article{edamadaka2025universally,
  title={Universally Converging Representations of Matter Across Scientific Foundation Models},
  author={Edamadaka, Sathya and Yang, Soojung and Li, Ju and G{\'o}mez-Bombarelli, Rafael},
  journal={arXiv preprint arXiv:2512.03750},
  year={2025}
}

@inproceedings{radford2021learning,
  title={Learning transferable visual models from natural language supervision},
  author={Radford, Alec and Kim, Jong Wook and Hallacy, Chris and Ramesh, Aditya and Goh, Gabriel and Agarwal, Sandhini and Sastry, Girish and Askell, Amanda and Mishkin, Pamela and Clark, Jack and others},
  booktitle={International conference on machine learning},
  pages={8748--8763},
  year={2021},
  organization={PmLR}
}

@article{facco2017estimating,
  title={Estimating the intrinsic dimension of datasets by a minimal neighborhood information},
  author={Facco, Elena and d’Errico, Maria and Rodriguez, Alex and Laio, Alessandro},
  journal={Scientific reports},
  volume={7},
  number={1},
  pages={12140},
  year={2017},
  publisher={Nature Publishing Group UK London}
}

@inproceedings{wang2020understanding,
  title={Understanding contrastive representation learning through alignment and uniformity on the hypersphere},
  author={Wang, Tongzhou and Isola, Phillip},
  booktitle={International conference on machine learning},
  pages={9929--9939},
  year={2020},
  organization={PMLR}
}

@article{diaconis1984asymptotics,
  title={Asymptotics of graphical projection pursuit},
  author={Diaconis, Persi and Freedman, David},
  journal={The annals of statistics},
  pages={793--815},
  year={1984},
  publisher={JSTOR}
}

@article{bac2021scikit ,
  title={Scikit-dimension: a python package for intrinsic dimension estimation},
  author={Bac, Jonathan and Mirkes, Evgeny M and Gorban, Alexander N and Tyukin, Ivan and Zinovyev, Andrei},
  journal={Entropy},
  volume={23},
  number={10},
  pages={1368},
  year={2021},
  publisher={MDPI}
}

@incollection{ghosh2009patchwork,
  title={Patchwork distributions},
  author={Ghosh, Soumyadip and Henderson, Shane G},
  booktitle={Advancing the frontiers of simulation: a Festschrift in honor of George Samuel Fishman},
  pages={65--86},
  year={2009},
  publisher={Springer}
}

@article{kulpa1999approximation,
  title={On approximation of copulas},
  author={Kulpa, Tomasz},
  journal={International Journal of Mathematics and Mathematical Sciences},
  volume={22},
  number={2},
  pages={259--269},
  year={1999}
}

@article{metropolis1953equation,
  title={Equation of state calculations by fast computing machines},
  author={Metropolis, Nicholas and Rosenbluth, Arianna W and Rosenbluth, Marshall N and Teller, Augusta H and Teller, Edward},
  journal={The journal of chemical physics},
  volume={21},
  number={6},
  pages={1087--1092},
  year={1953},
  publisher={American Institute of Physics}
}

@book{newman1999monte,
  title={Monte Carlo methods in statistical physics},
  author={Newman, Mark EJ and Barkema, Gerard T},
  year={1999},
  publisher={Clarendon Press}
}
\bibliographystyle{gram2026}
\newpage
\appendix
\etocdepthtag.toc{atoc} % Tag appendix sections as 'atoc'
\etocsettagdepth{mtoc}{none} % Ignore 'mtoc' (main text) in ToC
\etocsettagdepth{atoc}{subsection} % Show 'atoc' (appendix) down to subsections

% Optional: Style the ToC title (e.g., smaller font, specific name)
\etocsettocstyle{\subsection*{Appendix: Table of Contents}}{}

\tableofcontents % Print the ToC here
\section{Theoretical Framework}
\label{app:theory}

\subsection{Variational objectives and critic crchitectures}
\label{app:theory_objectives}

Our approach is based on variational estimation of mutual information (MI) between two random variables $X$ and $Y$ using contrastive objectives and neural critics. We briefly summarize the relevant objective and critic parameterizations used throughout the paper, and fix notation for the theoretical analysis that follows.

\paragraph{Variational MI estimators}
Mutual information can be written as a Kullback--Leibler divergence,
\begin{equation}
    I(X;Y) = D_{\mathrm{KL}}(p(x,y)\|p(x)p(y)) = \mathbb{E}_{p(x)} \left[ D_{KL}(p(y|x) \| p(y)) \right]. \label{eq:supp_mi_kullback}
\end{equation}

The Donsker-Varadhan~\citep{donsker1983asymptotic} representation provides a variational lower bound on the KL divergence between two distributions $P$ and $Q$:
\begin{equation}
    D_{KL}(P \| Q) \ge \sup_{T} \left( \mathbb{E}_{P}[T] - \log \mathbb{E}_{Q}[e^{T}] \right),
\end{equation}
where the supremum is taken over all measurable functions $T: \Omega \rightarrow \mathbb{R} $ on the common support, $\Omega$ of distributions $P$ and $Q$. Applying this to the various formulations of MI transforms the hard problem of calculating MI from estimating probabilities to optimizing over functions $T$. This is the basis of many MI estimation techniques, old and new \citep{nguyen2010estimating,Belghazi2018MutualEstimation,poole2019variational,song2019understanding}. 

Most neural-network-based MI estimation methods can be, directly or otherwise, derived from this formalism, with the most prominent estimators being MINE~\citep{Belghazi2018MutualEstimation}, SMILE~\citep{song2019understanding} and InfoNCE~\citep{oord2018representation}. We do not aim to provide a comprehensive review of these estimators in our work, with better reviews available elsewhere~\citep{poole2019variational}. We focus on the symmetrized-InfoNCE estimator, which in addition to being a variational MI estimator, has been widely used in the larger representation learning community under the guise of symmetrized contrastive losses or SimCLR objectives \citep{chen2020simple,radford2021learning}.

We stress, however, that our approach to dimensionality estimation is not dependent on the choice of estimator and works perfectly well with other estimators mentioned above as they all admit an optimal critic (or family of critics) very similar to the symmetrized-InfoNCE estimator. We have conducted experiments with other estimators including SMILE and MINE, which in general do not suffer from $\log$(batch size) bound of InfoNCE but suffer from larger variance with finite datasets~\citep{Belghazi2018MutualEstimation,song2019understanding}. For simplicity of discussion, we present results throughout with the symmetrized-InfoNCE estimator.

To derive the \textbf{symmetrized-InfoNCE} objective, we apply the DV representation to the conditional form in Eq.~(\ref{eq:supp_mi_kullback}), specifically to $D_{\rm KL}(p(y|x) \| p(y))$. We obtain:
\begin{equation}
    I(X;Y) \ge \sup_{T} \left( \mathbb{E}_{p(x,y)}[T(x,y)] - \mathbb{E}_{p(x)} \left[ \log \mathbb{E}_{p(y)}[e^{T(x,y)}] \right] \right).
\end{equation}
Similarly, applying this DV representation to $D_{\rm KL}(p(x|y) \| p(x))$, we obtain:
\begin{equation}
    I(X;Y) \ge \sup_{{T'}} \left( \mathbb{E}_{p(x,y)}[{T'}(x,y)] - \mathbb{E}_{p(y)} \left[ \log \mathbb{E}_{p(x)}[e^{{T'}(x,y)}] \right] \right),
\end{equation}
where $T'$ also independently consists of all measurable functions from the common support of $X$ and $Y$.

Of course, this implies
\begin{equation}
\begin{aligned}
    I(X;Y) &\ge \frac{1}{2}\Big\{\sup_{T} \left( \mathbb{E}_{p(x,y)}[T(x,y)] - \mathbb{E}_{p(x)} \left[ \log \mathbb{E}_{p(y)}[e^{T(x,y)}] \right] \right) \\
    &\quad\quad  + \sup_{{T'}} \left( \mathbb{E}_{p(x,y)}[{T'}(x,y)] - \mathbb{E}_{p(y)} \left[ \log \mathbb{E}_{p(x)}[e^{{T'}(x,y)}] \right] \right)\Big\}\\
    &\ge\sup_T \left( \mathbb{E}_{p(x,y)}[T(x,y)] - \frac{1}{2}\left( \mathbb{E}_{p(x)} \left[ \log \mathbb{E}_{p(y)}[e^{T(x,y)}] \right] + \mathbb{E}_{p(y)} \left[ \log \mathbb{E}_{p(x)}[e^{{T}(x,y)}] \right] \right) \right) \\
    &:= \mathcal{I}_{\text{symm-NCE}}\left(T\right)
\end{aligned}
\end{equation}

From above, it is clear that $\mathcal{I}_{\text{symm-NCE}}\left(T\right)$ is bounded above by true MI. To show that it serves as a variational estimator, it suffices to show that there exists an optimal critic $T^*$ such that $\mathcal{I}_{\text{symm-NCE}}\left(T^*\right) = I$. This can be easily verified  for $T^* = \log \dfrac{p(x, y)}{p(x) p(y)} +\text{c}$. This is indeed the same optimal critic as for other variational estimators such as MINE~\citep{poole2019variational}.

In practice, the expectations are replaced by using contrastive sampling within a batch. For a batch $\{(x_i, y_i)\}_{i=1}^N$, we treat $y_i$ as the positive sample for $x_i$ and the other $N-1$ samples $\{y_j\}_{j \ne i}$ as negative samples from the marginal $p(y)$ \citep{oord2018representation}. This yields the estimator:
\begin{equation}
    I_{\text{symm-NCE}}(X,Y) := \frac{1}{2N} \left(\sum_{i=1}^{N} \log \frac{e^{T(x_i, y_i)}}{\frac{1}{N} \sum_{j=1}^{N} e^{T(x_i, y_j)}} + \sum_{j=1}^{N} \log \frac{e^{T(x_j, y_j)}}{\frac{1}{N} \sum_{i=1}^{N} e^{T(x_i, y_j)}} \right).
\end{equation}
This estimator is low-variance but bounded above by $\log N$.

\paragraph{Critic architectures}

The MI estimators turn the MI estimation problem  into an optimization problem of the variational bound, but the structure of the learned representation is controlled by the design of the critic function $T(x,y)$. 

The critic determines how correlations between $X$ and $Y$ are represented. While sufficiently expressive critics can approximate the optimal solution of the variational problem by representing the optimal critic $T^*$, their internal structure governs whether correlations are able to be decomposed into low-dimensional degrees of freedom. This distinction plays a central role in our ability to extract an effective dimensionality from the representation optimized for MI estimation.

A common choice in neural MI estimation is a \emph{concatenated} or joint critic~\citep{Belghazi2018MutualEstimation}, in which $X$ and $Y$ are combined by a generic function,
\begin{equation}
    T_{\text{concat}}(x,y) = T_\theta([x,y]),
\end{equation}
where $T_\theta$ is typically a MLP. This parameterization places minimal structural constraints on how correlations are represented. As a result, dependencies between $X$ and $Y$ can be distributed across the full joint representation, with no explicit notion of ordered degrees of freedom. While these critics are well suited for MI estimation, they do not expose a notion of intrinsic or effective dimensionality, and they are often more expensive to train, compared to the separable critic for example.

In contrast, another common design choice is \emph{separable} critic architectures~\citep{oord2018representation}, which decompose the interaction between $X$ and $Y$ into a sum of factorized terms,
\begin{equation}
    T_{\text{sep}}(x,y) = g^X(x) \cdot g^Y(y) = \sum_{k=1}^{k_z} g^X_k(x) g^Y_k(y).
    \label{eq:separable_critic}
\end{equation}
Here, the critic represents a scalar product between embeddings of the two datasets via the \emph{encoders} $g^X : \mathcal{X} \to \mathbb{R}^{k_z}$ and
$g^Y : \mathcal{Y} \to \mathbb{R}^{k_z}$. This form explicitly enforces a pairing structure between learned features of the two variables and decomposed into (at most) $k_z$ modes. 

Separable and concatenated critics impose qualitatively different constraints on how correlations
are represented, but neither is naturally useful for measuring the latent dimensionality: separable critics restrict
interactions to a  bilinear form at a prescribed embedding dimension\footnote{The relevance of latent dimensionality for NN-based MI estimators is becoming more popular (see, e.g, ~\cite{gowri2024approximating}), but such work has relied on using existing separable architectures, which, as we have shown comes with stark limitations and inflated dimensionality estimation.}, while concatenated
critics are flexible but can entangle dependencies across coordinates, obscuring their organization.

In this work, we bridge the gap and motivate a \emph{\textbf{hybrid}} architecture that can force accurate low-dimensional, latent space representations and then optimize over a flexible concatenated head to represent the optimal critic in the learned latent space. As we demonstrate throughout the paper, this separation allows the encoders to faithfully represent the latent degrees of freedom, while the concatenated head optimizes the variational objective without imposing additional structural constraints. The effective dimensionality is therefore reflected in the learned embeddings themselves:
\begin{equation}
    T_{\text{hybrid}}(x, y) = T_\theta ([g^X(x), g^Y(y)]).
\end{equation}

In principle, the choice of critic architecture can be entirely decoupled from the choice of the variational estimator (i.e., objective). In this paper, we show that the \emph{hybrid} architecture, coupled with the symmetrized-InfoNCE, not only can accurately estimate MI, sometimes outperforming existing estimators in the number of samples required (Fig.~\ref{fig:si_gauss_mix_MI_samples}), but also reveals the correct latent space dimensionality across a wide variety of latent spaces and distributions.

\subsection{Exactly solvable case: Jointly Gaussian variables}
\label{app:theory_solvable}
In the case of jointly Gaussian variables, the optimal critic $T^*$ is known analytically. By examining the functional form of the optimal critic in this simple setting, we can validate our numerical results, demonstrate the limitations of a separable architecture and highlight the utility of the \emph{hybrid} design.

\subsubsection{Derivation of the Optimal Gaussian Critic}
\label{app:theory_gaussian_derivation}

This derivation largely follows the pedagogical introduction in \cite{abdelaleem2025accurate}.
Consider $X \in \mathbb{R}^{K_X}$ and $Y \in \mathbb{R}^{K_Y}$ distributed as a zero-mean jointly Gaussian variable $Z = [X, Y]^\top$:
\begin{equation}
    Z \sim \mathcal{N}\left(0, \Sigma \right), \quad \Sigma = \begin{bmatrix} \Sigma_{XX} & \Sigma_{XY} \\ \Sigma_{YX} & \Sigma_{YY} \end{bmatrix}.
\end{equation}

The optimal critic (for the population-level variational objective underlying the symmetrized-InfoNCE estimator), i.e., $\mathcal{I}_{\text{symm-NCE}}(T^*)  =I(X; Y)$, is given by:
\begin{equation}
    T^*(x,y) = \log \frac{p(x,y)}{p(x)p(y)} + C.
\end{equation}

Substituting the probability density functions for the multivariate Gaussian distribution,
\begin{equation}
    p(x, y) \propto \exp\left( -\dfrac{1}{2}\begin{bmatrix}
        x\\y
    \end{bmatrix}^\top \Sigma^{-1}\begin{bmatrix}
        x \\y
    \end{bmatrix} \right), \;\;\;\; p(x) p(y) \sim \exp\left(-\frac{1}{2} x^\top \Sigma_{XX}^{-1} x -\frac{1}{2}y^\top \Sigma_{YY}^{-1} y \right), 
\end{equation}

yields a quadratic form for the optimal critic
\begin{equation}
    T^*(x,y) = \frac{1}{2} \left[ x^\top \Sigma_{XX}^{-1} x + y^\top \Sigma_{YY}^{-1} y - \begin{bmatrix} x \\ y \end{bmatrix}^\top \Sigma^{-1} \begin{bmatrix} x \\ y \end{bmatrix} \right] + C.
    \label{eq:optimal_quadratic}
\end{equation}

This explicitly shows that the optimal critic for jointly Gaussian data is a quadratic function of the inputs. To analyze this, it is useful to transform to a canonical coordinate system.

To do so, define the whitened (i.e., normalized) cross-covariance matrix, $\mathcal{K} = \Sigma_{XX}^{-1/2} \Sigma_{XY} \Sigma_{YY}^{-1/2}$ and then perform its singular value decomposition, $\mathcal{K} = U \Lambda V^\top$, where $\Lambda = \text{diag}(\rho_1, \dots, \rho_{r})$, with $r={\rm rank\,} {\mathcal K}$. We can define the canonical variables $u$ and $v$ via linear projections
\begin{equation}
    u = U^\top \Sigma_{XX}^{-1/2} x, \;\;\; v = V^\top \Sigma_{YY}^{-1/2} y.
\end{equation}

In this coordinate system, the joint covariance matrix $\Sigma$ simplifies into blocks for each canonical pair $(u_i, v_i)$ with the inverse covariance matrix $\Sigma^{-1}$ for the $i$-th pair given by:
\begin{equation}
    (\Sigma^{(i)})^{-1} = \frac{1}{1 - \rho_i^2} \begin{bmatrix} 1 & -\rho_i \\ -\rho_i & 1 \end{bmatrix} .
\end{equation}

Using this, the optimal critic can be written as a sum over independent canonical coordinates, decomposed into bilinear and quadratic terms:
\begin{equation}
    T^*(x,y) = \sum_{i=1}^{r} \left[ \frac{\rho_i}{1 - \rho_i^2} u_i v_i - \frac{1}{2} \frac{\rho_i^2}{1 - \rho_i^2} (u_i^2 + v_i^2) \right].
    \label{eq:optimal_critic_expanded}
\end{equation}

The first interaction term is purely \emph{bilinear} in ($u_i v_i$), while the \emph{self-normalization} terms are quadratic in $u_i$ and $v_i$. This derivation effectively recovers the standard Canonical Correlation Analysis (CCA) objective~\citep{kullback1959information,gelfand1959calculation}. Below, we use this expression to design a quadratic critic architecture (\emph{separable-augmented}) for the jointly Gaussian case, demonstrating how it indeed can be used to infer the dimensionality of the latent distribution, as well as explore the role of embedding dimension in separable critics. 

\subsubsection{Choice of architecture and optimal embeddings}
\label{app:theory_separable_augmented}

Given the  optimal critic, Eq.~(\ref{eq:optimal_critic_expanded}), it is clear that a $K$ dimensional jointly Gaussian distribution is not in the class of separable critics with encodings in $k_z = K$ dimensions (here and throughout this section we simplify notation with $K$ in place of $K_Z$ for the latent dimensionality). In particular, the ``self-normalization'' terms cannot be represented as a dot product in $K$-dimensional space. 

However, this is typically not a limitation for MI estimation using separable architectures, where the encoders are non-linear and typically embed in a much larger space than the true latent structures. To see that a bilinear representation of an optimal critic is possible with non-linear encoders in $K+2$ dimensions, we can directly check that $T(x, y) = g^X \cdot g^Y = T^*(x, y)$ for the following  encoding:
\begin{equation}
\begin{aligned}
    g^X(x) &= \left(\sqrt{\frac{\rho_1}{1-\rho_1^2}}u_1, \cdots, \sqrt{\frac{\rho_K}{1-\rho_K^2}} u_K, \sum_i^K -\frac{\rho_i^2}{2(1-\rho_i^2)} u_i^2, 1\right) \in \mathbb{R}^{K+2}, \\
    g^Y(y) &= \left(\sqrt{\frac{\rho_1}{1-\rho_1^2}}v_1, \cdots, \sqrt{\frac{\rho_K}{1-\rho_K^2}} v_K, 1, \sum_i^K -\frac{\rho_i^2}{2(1- \rho_i^2)} v_i^2\right) \in \mathbb{R}^{K+2}.\label{eq:supp_k_2_encoders}
\end{aligned}
\end{equation}
This is one of many degenerate solutions that can represent the optimal critic via a dot product in $K+2$ dimensions. The scalar product in the first $K$ dimensions captures the interaction terms in Eq.~(\ref{eq:optimal_critic_expanded}), and the two extra dimensions  individually capture the normalization terms for $X$ and $Y$.

Note that this is only for estimators (MINE, SMILE, symmetrized-InfoNCE) where the optimal critic corresponds to $T^* = \log \left( p(x, y)/p(x)p(y)\right) + C$. With nonsymmetrized InfoNCE, the optimal critic is given by $T^* = \log p(y|x) + \tilde{c}(y) =  \log \left(p(x, y)/p(x)p(y)\right) + c(y)$~\citep{poole2019variational}, with the functional degeneracy represented in $c(y)$. In this case, with the separable architecture the optimal critic family can clearly be encoded in $K+1$ dimensions, by, e.g., the first $K+1$ dimensions of the encoders given in Eq.~(\ref{eq:supp_k_2_encoders}) and discarding the normalization terms for $y$. 

\begin{figure}[t]
    \centering
    \includegraphics[width=0.8\linewidth]{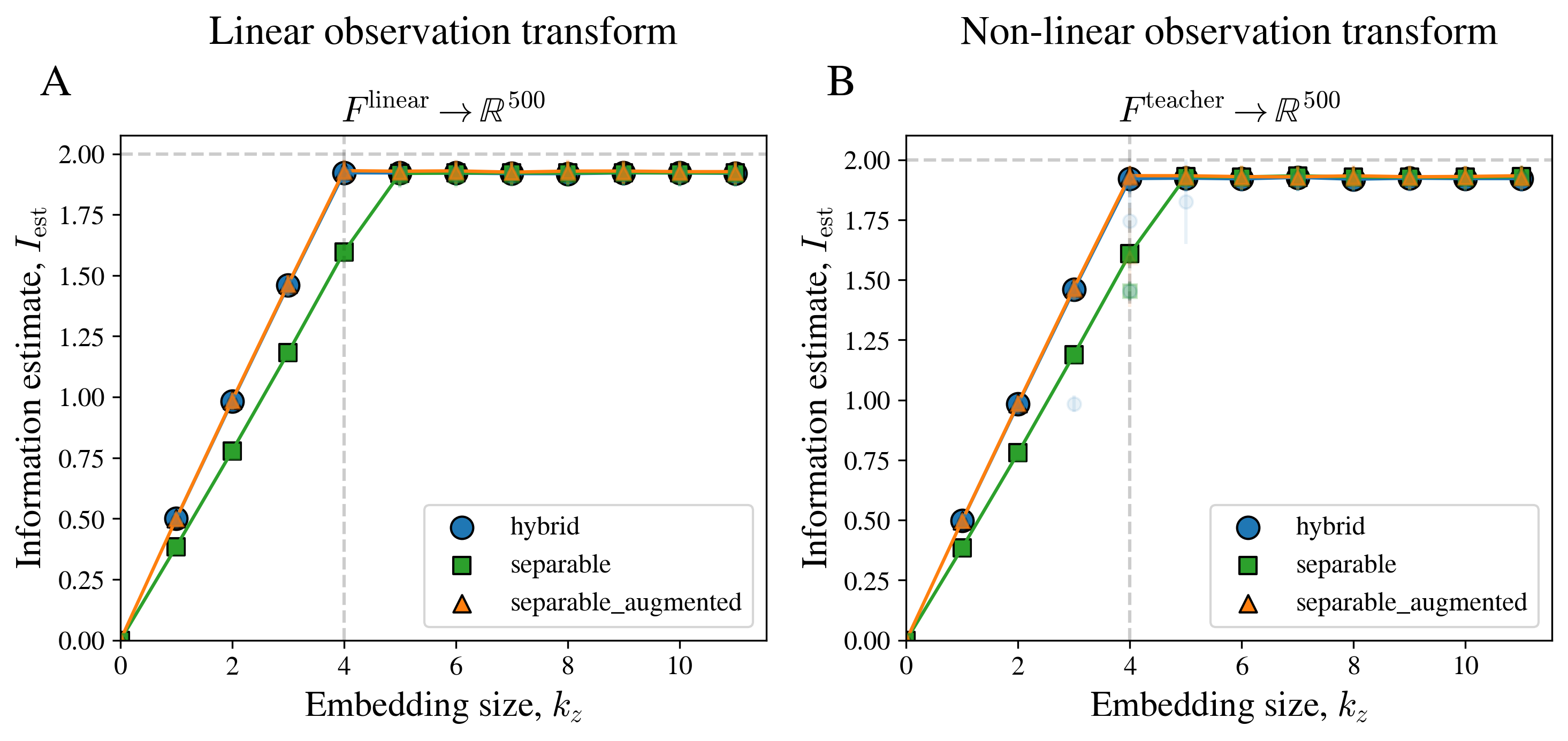}
    \caption{\textbf{Required embedding dimensionality for different critics.} The separable augmented critic, with the \emph{symmetrized-InfoNCE} estimator also recover the true MI, in embedding dimension $k_z = K_Z$ for a joint Gaussian distribution. The figures supplements Fig.~\ref{fig:fig2_infinite_synthetic_joint} in the main text, for the joint Gaussian distribution with $K_Z =4$, passed through linear and non-linear teacher transforms to $\mathbb{R}^{500}$ to create the observed data. Individual runs shown by semi-transparent markers. }
    \label{fig:supp_sep_augmented}
\end{figure}

To validate that the correct MI estimation (with a known latent distribution) corresponds to the critic learning the correct latent representation Eq.~(\ref{eq:optimal_critic_expanded}), we construct an extended quadratic family of critics that can be learned while training, and we explore whether such an architecture allows for the optimal critic to be learned in $K$ dimensions exactly. Using encoders $g^X(x)$ and $g^Y(y)$, we define:
\begin{equation}
    T_{\text{separable-augmented}}(x, y) = (g^X)^{\top} g^Y + (g^X)^{\top} \gamma^X g^X + (g^Y)^{\top} \gamma^Y g^Y   
\end{equation}
where, $\gamma^X, \gamma^Y$ are ($k_z \times k_z$) weight matrices composed of trainable parameters. With correctly learnt encoders and $\gamma^X, \gamma^Y$ it is obvious that the optimal critic, Eq.~(\ref{eq:optimal_critic_expanded}), can be represented by the \emph{separable-augmented} critics in $k_z =K$ dimensions. This is indeed what we find, as shown in Fig.~\ref{fig:supp_sep_augmented}.

A curious observation in these experiments is that even with the \emph{separable} critics and symmetrized-InfoNCE estimator such that the optimal critic $T^*(x, y) = \log\left(p(x, y)/p(x)p(y)\right)$, we empirically find that $K+1$ dimensions suffice to learn the true MI (and not $K+2$ as we argued above would, at most, suffice). 
There is no obvious solution in $K+1$ dimensions that would actually be sufficient to represent this optimal critic. Visualizing the learnt encoders (with $k_z = K+1$) in our experiments with the joint Gaussian distributions, we observe spherical embeddings (not shown). This leads us to consider an interesting ansatz: encoding on hypersphere $S^K$, embedded in $\mathbb{R}^{K+1},$ can lead to accurate MI for a $K$-dimensional joint Gaussian with separable critics.

To verify this, we first restrict  to $K=1,$ i.e., 1-dimensional joint Gaussians, with embeddings: 
\begin{equation}
\begin{aligned}
    g^X(x) &= \left(\alpha_1 \cos (u/\beta_1), \alpha_2 \sin(u/\beta_1)\right),  \\
    g^Y(y) &= \left(\alpha_3 \cos (v/\beta_2), \alpha_4 \sin(v/\beta_2)\right),
\end{aligned}
\end{equation}
where, as before, $u, v$ are the canonical coordinates corresponding to $x, y$ respectively such that $\mathbb{E}[u] =\mathbb{E}[v] =0, $ $\mathbb{E}[u^2] =\mathbb{E}[v^2] =1$. The correlation $\rho$ is defined as $\rho = \mathbb{E}[uv]$ and then the true MI is $I = -\dfrac{1}{2}\log_2 (1-\rho^2)$. Here, $\alpha_i, \beta_i$ are parameters we will use to optimize the MI estimate from this embedding.

The separable critic is then given by
\begin{equation}
\begin{aligned}
    T(x,y) &= g^X \cdot g^Y = \alpha_1 \alpha_3 \cos(u/\beta_1)\cos(v/\beta_2)  + \alpha_2\alpha_4\sin(u/\beta_1)\sin(v/\beta_2)\\
    &= \lambda_+ \cos(au + bv) + \lambda_- \cos(au - bv),
\end{aligned}
\end{equation}
with $\lambda_{\pm} = \frac{1}{2}(\alpha_1\alpha_3 \mp \alpha_2\alpha_4)$. Here, for tractability, we restrict ourselves to $\lambda_+=0$, which can only provide a lower bound on the estimated MI. As we will show, even with this restricted subspace, we can approximate the true MI very well. And this restriction corresponds to $\alpha_1=\alpha_2, \alpha_3 =\alpha_4$ i.e. a circular embedding. So, we optimize the symmetrized-InfoNCE objective over $\lambda, a,b$ where
\begin{equation}
    T(x,y) = \lambda \cos(au - bv).
\end{equation}

With this restricted critic, to optimize the symmetrized-InfoNCE estimate over the critic means to optimize the bound:

\begin{equation}
\begin{aligned}
    \mathcal{I}(a,b, \lambda; \rho) = \mathbb{E}[T] - \frac{1}{2} \mathbb{E}_X\log \mathbb{E}_{Y} e^T
    - \frac{1}{2} \mathbb{E}_Y\log \mathbb{E}_{X}e^T \label{eq:supp_ical_abr}
\end{aligned}
\end{equation}

such that 
\begin{equation}
    \mathcal{I}^{\text{circle}}(\rho) \equiv \sup_{a, b, \lambda}\; \mathcal{I}(a,b, \lambda; \rho). 
\end{equation}

To compute this optimal estimate, $\mathcal{I}^{\text{circle}}$, we make use of two identities:
\begin{enumerate}
    \item 
For $W\sim \mathcal{N}(\mu, \sigma^2)$
\begin{equation}
    \mathbb{E}[e^{inW}] = e^{in\mu}e^{-\frac{1}{2} n^2\sigma^2}.
\end{equation}
Which directly implies, $\mathbb{E}[\cos(nW)] = \cos(n\mu) e^{-n^2\sigma^2/2}.$

\item
To calculate the expectation over the exponential of a sinusoid $\mathbb{E}[e^{\lambda \cos(au- bv)}]$, we  use the Jacobi expansion using Bessel functions and then compute the expectations as above with
\begin{equation}
    e^{z\cos\theta} =  I_0(z) + 2\sum_{n=1}^\infty I_n(z) \cos(n\theta),
\end{equation}
where $I_n$ are modified Bessel functions of the first kind.
\end{enumerate}

With these identities, we can write down the various expectations needed for the estimator in Eq.~(\ref{eq:supp_ical_abr}). The first term is given by:
\begin{equation}
\begin{aligned}
    \mathbb{E}[T] = \lambda \exp \left[{-\frac{1}{2}(a^2+b^2 -2ab \rho)}\right]\;. \\
\end{aligned}
\end{equation}

For the $\log$ terms, we cannot compute the expectations analytically, but we can reduce it to quadrature, as below, and then numerically optimize over $a,b,$ and $\lambda$. This gives:
\begin{equation}
\begin{aligned}
\mathbb{E}_{Y} e^T &= \mathbb{E}_{Y}\left[e^{\lambda \cos (au - bv)}\right]
\\&=\mathbb{E}_{Y}\left[ I_0(\lambda) + 2\sum_{n=1}^\infty I_n(\lambda) \cos(n(au - bv ))\right]\\
&=I_0(\lambda) + 2\sum_{n=1}^\infty I_n(\lambda) \cos(nau)e^{-n^2b^2/2}.
\end{aligned}
\end{equation}
And then we can compute $\mathbb{E}_X {\log(\mathbb{E}_Ye^T)} $ numerically.

From symmetry of $a\rightleftharpoons b$ in the full objective function, (given a unique optimizer) the optimum will lie at $a=b :=  \kappa$ and then both the row and column terms are equal, such that
\begin{equation}
    \mathcal{I(\kappa, \lambda; \rho)} = \lambda e^{-\kappa^2(1-\rho)} - \mathbb{E}_Z\left[\log\left(I_0(\lambda) + 2\sum_{n=1}^\infty I_n (\lambda) \cos(n \kappa z) e^{-n^2\kappa^2/2}\right)\right],
    \label{eq:supp_circle_estimator}
\end{equation}
where $z\in Z\sim \mathcal{N}(0,1),$ $I_n$ are the modified Bessel functions of the first kind, and $
I^{\text{circle}}(\rho) \ge \max_{\kappa,\lambda}\; \mathcal{I}(\kappa, \lambda; \rho).$

\begin{figure}[t]
    \centering
    \includegraphics[width=0.8\linewidth]{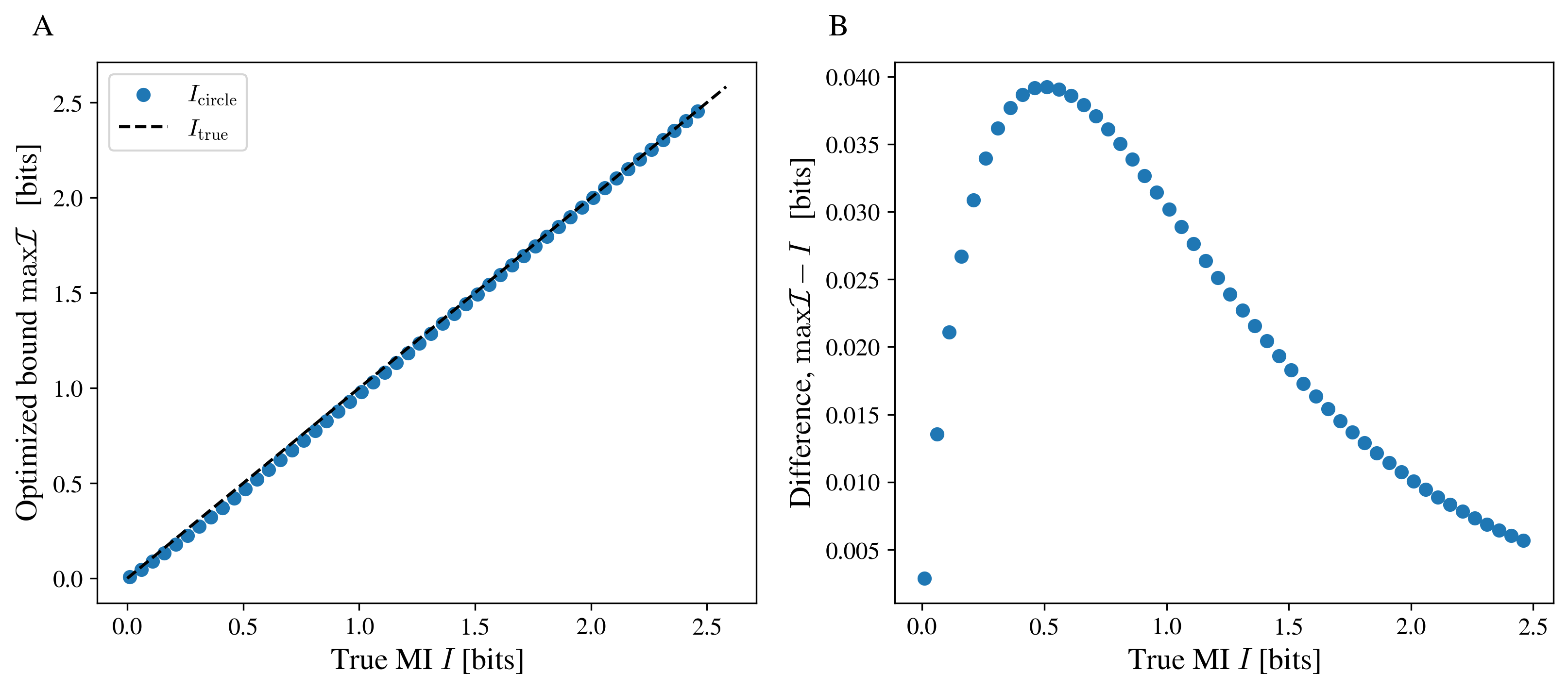}
    \caption{\textbf{Bilinear critic produces circular embeddings.} We estimate the variational bound for the circular embedding by numerically approximating $\mathcal{I}(\kappa, \lambda|\rho)$ Eq.~(\ref{eq:supp_circle_estimator}) using quadratures and then optimizing over a grid of $\kappa, \lambda$ to get the variational bound $\mathcal{I}^{\text{circle}}(\rho)$. As discussed in the text, we find that this estimator can approach the true MI, with a gap $< 0.04$ bits, showing how a two dimensional separable critic can approach the true MI for a one-dimensional joint Gaussian distribution.}
    \label{fig:supp_circle_estimator}
\end{figure}

We approximate this expectation over Gaussian $z$ numerically using Gauss-Hermite polynomials, and as shown in Fig.~\ref{fig:supp_circle_estimator}, we find that $\max |I_{\text{circle}} - I| \approx 0.04 $ bits, with the peak at $I \approx 0.5$ bits or equivalently $\rho = \sqrt{1-2^{-2I}} \approx 0.7$.
Such a gap would be well within the fluctuations (due to finite data/batch sizes) in our experiments with the separable critics and, therefore, would register as a saturation of the MI estimate for $k_z = K+1$. 

Note, we have not shown this to be true for $K>1$. For the joint Gaussian, with equal correlations in $K$ directions, as has been considered throughout this paper, a natural extension of this circular embedding is to embed on an $S^K$-sphere in $\mathbb{R}^{K+1}$, i.e., with angles $\theta_i = \kappa_i u_i$ such that the encoder $g^X$ has components $g_1 = \lambda \cos \theta_1$, $g_2 = \lambda \sin \theta_1 \cos\theta_2$,  $\cdots,$ $g_K =\lambda \sin \theta_{K-1} \cdots \sin \theta_1 \cos \theta_K$, $g_{K+1} = \lambda\sin \theta_K \cdots \sin \theta_1$, and analogously for $g^Y$. With the same symmetry constraints as in the 1-d case with $\kappa \equiv \kappa^X_i = \kappa^Y_i$, and using a more general expansion of the $d$-dimensional sinusoidal products, one can then consider a more general expansion of the exponential expectations.

While we do not pursue this calculation further here, and consider this beyond the scope/relevance of this work, our numerical experiments with separable critics and the symmetrized-InfoNCE estimator strongly suggest that such hyperspherical embeddings are sufficient to achieve near-saturation of MI bound with embedding dimension $k_z = K+1$. This observation is consistent with general empirical findings in representation learning, where contrastive losses (which as we showed earlier, are equivalent to MI optimization) and separable architectures (e.g., SimCLR) tend to produce approximately uniform and aligned embeddings on the hypersphere in sufficiently high-dimensional latent spaces~\citep{wang2020understanding}. In sufficiently high-dimensional latent spaces, a wide class of non-pathological distributions exhibit approximately Gaussian statistics along most directions, essentially due to aggregation of independent factors~\citep{diaconis1984asymptotics}, and, hence, the optimal embedding which maximizes the variational MI objectives, using separable architectures, would result in hyperspherical embeddings.

\section{Design choices: Estimation and Stopping Protocols and Effective Dimensionality Measures}
\label{app:protocol}

\subsection{Max-Test / Train-Estimate protocol}
\label{app:protocol_maxtest}

A fundamental challenge that is often overlooked in neural-network-based MI estimators when training on finite datasets is determining when to stop training. Unlike supervised learning, where a test loss plateau indicates convergence, variational MI lower bounds can grow indefinitely as the critic overfits to finite-sample artifacts. For example, InfoNCE can keep growing until saturation at $\log N$ while SMILE can keep growing past that point.

To resolve this, we follow \cite{abdelaleem2025accurate} and employ a \textbf{Max-Test / Train-Estimate} protocol. We split the available data into a training set ($\mathcal{D}_{\text{train}}$) and a test set ($\mathcal{D}_{\text{test}}$). At each training epoch $t$, we evaluate the estimator on both sets:
\begin{align}
    \hat{I}_{\text{train}}^{(t)} &= \mathcal{L}_{\text{EST}}(\mathcal{D}_{\text{train}}, \theta_t), \\
    \hat{I}_{\text{test}}^{(t)} &= \mathcal{L}_{\text{EST}}(\mathcal{D}_{\text{test}}, \theta_t).
\end{align}
We select the optimal stopping point $t^*$ as the epoch that maximizes the test estimate: $ t^*=\arg\max_t \hat{I}_{\text{test}}^{(t)}$. However, crucially, we report $\hat{I}_{\text{train}}^{(t^*)}$ as the final estimate, rather than the test value.\footnote{Several practical choices can reduce overhead. One can monitor a proxy for
$\hat I_{\text{train}}$ by evaluating on a subset of the training data rather than the full set. One
can also avoid computing $\hat I_{\text{train}}$ at every epoch: evaluate only the test estimate,
save the checkpoint whenever $\hat I_{\text{test}}$ improves, and at the end compute
$\hat I_{\text{train}}^{(t^\ast)}$ once on the full training set using the best checkpoint. Note
that for dimensionality estimation knowing  the MI value per se is not crucial  (e.g., in settings like
$I(g^X(Z);g^Y(Z))$), but it is still useful to track MI to detect estimator saturation (e.g., the
$\log N$ ceiling for InfoNCE) and to decide whether to switch estimators, stop earlier, increase the
batch size, etc. Finally, runs that fail to learn any nontrivial MI should be discarded.}

\subsection{Effective dimensionality}
\label{app:protocol_pr}

To quantify the effective dimensionality of the learned representations without performing an exhaustive sweep over embedding sizes $k_z$, we analyze the spectrum of the learned embeddings. We train a single \emph{hybrid} estimator with an overparameterized bottleneck ($k_z \gg K_{Z}$).

Let $Z_X = g^X(X)$ and $Z_Y = g^Y(Y)$ be the batch of embeddings produced by the encoders. We compute the centered cross-covariance matrix:
\begin{equation}
    C_{XY} = \frac{1}{N-1} (Z_X - \bar{Z}_X)^\top (Z_Y - \bar{Z}_Y).
\end{equation}
We then compute the singular values $\{\sigma_i\}$ of $C_{XY}$. 
The effective dimensionality $d_{\text{eff}}$ is derived from this spectrum. The choice of metric for $d_{\text{eff}}$ involves a choice of how strongly small values in the spectrum are counted or what constitutes a gap in the spectrum, which resembles how do we decide that we see a saturation in $I$ vs $k_z$ curves in case of absence of a clear sharp knee in the curve.

In this paper, we define $d_{\rm eff}$ via the Participation Ratio (PR) of the singular values:
\begin{equation}
    d_{\text{eff}} = \frac{(\sum_i \sigma_i)^2}{\sum_i \sigma_i^2}.
    \label{eq:pr_sv_definition}
\end{equation}

This choice of metric corresponds to a choice of how to filter the learned spectrum into an effective scalar dimensionality estimate. This could be replaced with different choices: we can either threshold the (normalized) singular value spectrum to count the number of non-trivial modes, or define various continuous analogues to the participation ratio, e.g.: 
\begin{enumerate}
    \item {PR based on Eigenvalues:} Similar to Eq.~(\ref{eq:pr_sv_definition}), we can define: 
    \begin{equation}
        d^{\rm{eig}}_{\text{eff}}=\frac{(\sum_i \lambda_i)^2}{\sum_i \lambda_i^2} = \frac{(\sum_i \sigma_i^2)^2}{\sum_i \sigma_i^4}.
        \label{eq:pr_eig_definition}
    \end{equation}Such definition kills smaller values of the spectrum faster, yielding a smaller estimate for $d_\text{eff}$.
    \item {PR based on Spectral Entropy:} Defining dimensionality as the exponential of the spectral entropy, $e^{H(\tilde{\sigma})}$, where $\tilde{\sigma}_i = \sigma_i/\sum_i \sigma_i$ are normalized singular values: 
    \begin{equation}
        d^{\rm{s-ent}}_\text{eff} = \exp{\{-\sum_i\tilde{\sigma}_i\log{\tilde{\sigma}_i}\}}.
        \label{eq:pr_entropy_definition}
    \end{equation}
    Such a definition often inflates the dimensionality estimate by including smaller components, leading to estimates of $d_{\text{eff}} > K_{\text{true}}$ in noisy regimes.
    \item {Intermediate PR}:
    \begin{equation}
        d_{\rm{eff}}^\alpha = \frac{\left(\sum \sigma_i^\alpha\right)^2}{\sum \sigma_i^{2\alpha}}
    \end{equation}
\end{enumerate}

\begin{figure}[t]
    \centering
    \includegraphics[width=0.45\linewidth]{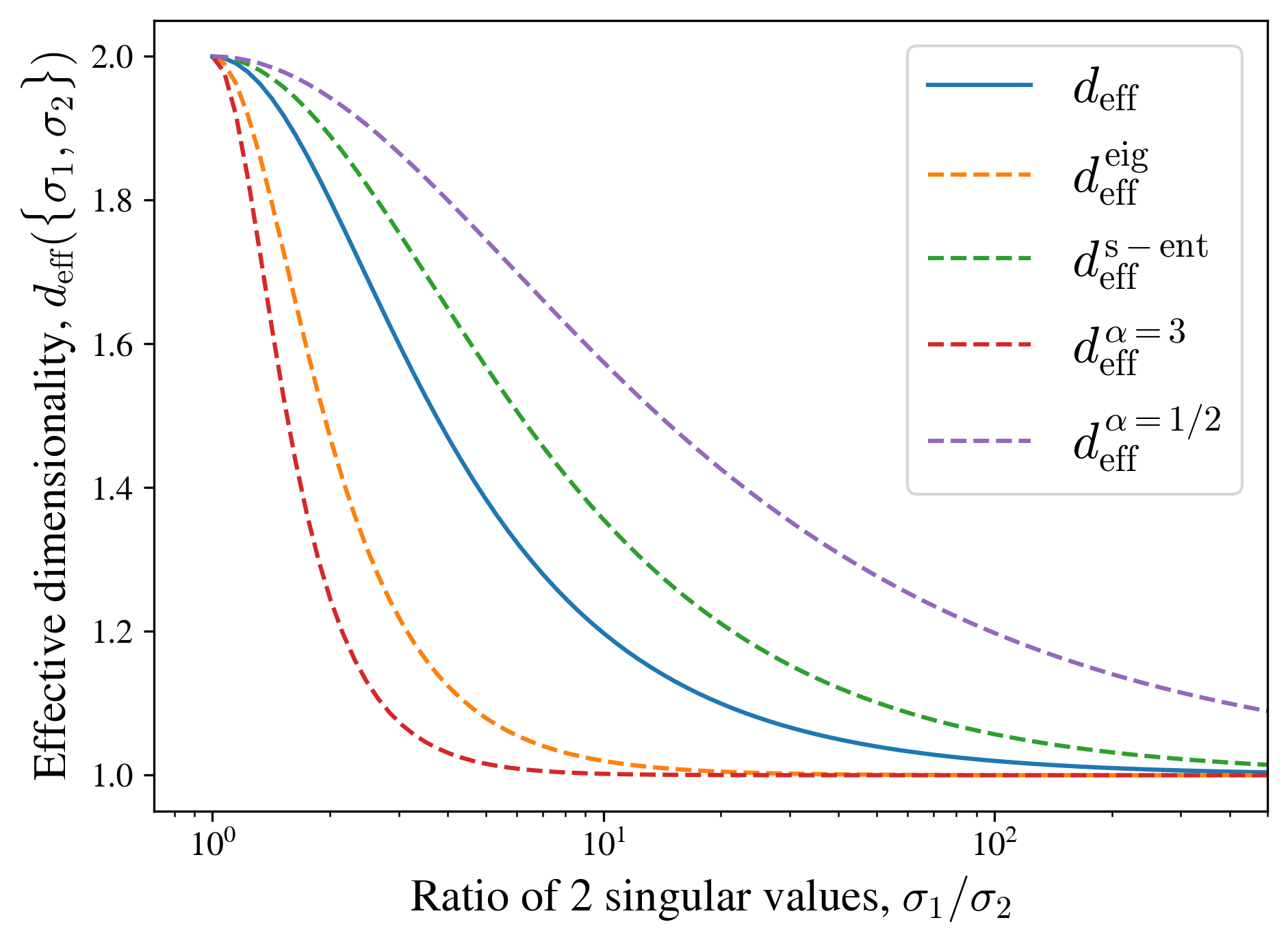}
    \caption{\textbf{Effective dimensionality based on the various metrics for a pair of singular values, $\sigma_1$ and $\sigma_2$, as a function of their ratio.} With both singular values equal, the effective dimensionality is 2. As they become unequal, the effective dimensionality becomes smaller, with the rate of decrease determined by the choice of the metric. Throughout this paper, we have used the participation ratio of the singular values (shown here with the solid line) of the cross-covariance of the encoders as the measure for effective dimensionality.}
    \label{fig:supp_deff_definitions}
\end{figure}

As illustrated in Fig.~\ref{fig:supp_deff_definitions}, various choices of such a metric relate to how to consider contributions from unequally dominant modes in the embeddings, and, in principle, should depend on the physical system under consideration. For concreteness, throughout this paper, we default to the PR of the singular values as our metric of choice as it provides a robust middle ground, but there is no a priori reason for such a choice. For spectra with equally dominant modes, all the choices would converge and provide very similar estimates of effective dimensionality.

% ==============================================================================
% APPENDIX C:
% ==============================================================================
\section{Dimensionality estimation with observation noise}
\subsection{View splitting  for intrinsic dimensionality estimation}
\label{app:shared_latent}

In Sec.~\ref{sec:results-synthetic}, we validated our estimator on synthetic datasets where distinct
views $X$ and $Y$ were explicitly generated. However, in scientific applications
(Sec.~\ref{subsec:results-real}), we often possess a single high-dimensional dataset (e.g., a video
or a spin configuration) and wish to infer its intrinsic dimensionality. Here, we provide the
theoretical and empirical justification for applying our estimator in this ``single-dataset''
regime. We model the data as arising from a \emph{Shared Latent Space} model:
\begin{equation}
    X = F_X(Z) + \eta_X, \quad Y = F_Y(Z) + \eta_Y,
    \label{eq:noise_view_splitting}
\end{equation}
where $Z$ is the underlying latent variable (drawn from $p_Z$), $F_X, F_Y$ are (potentially
identical) observation maps, and $\eta_X, \eta_Y$ are uncorrelated noise terms.

Crucially, the presence of observation noise is not a hindrance but a requirement for dimensionality
estimation via MI. In the absence of noise ($\eta \to 0$), if $F$ is deterministic and invertible,
MI would diverge (or, more precisely, equal the entropy $H(Z)$, which is parameterization
dependent), making it difficult to define a saturation point relative to the embedding dimension.
With uncorrelated noise, the information bottleneck is determined by the shared signal $Z$, and one
expects the estimated MI to saturate when the embedding dimension $k_z$ matches $\dim(Z)$.

\begin{figure}[t]
    \centering
    \includegraphics[width=0.7\linewidth]{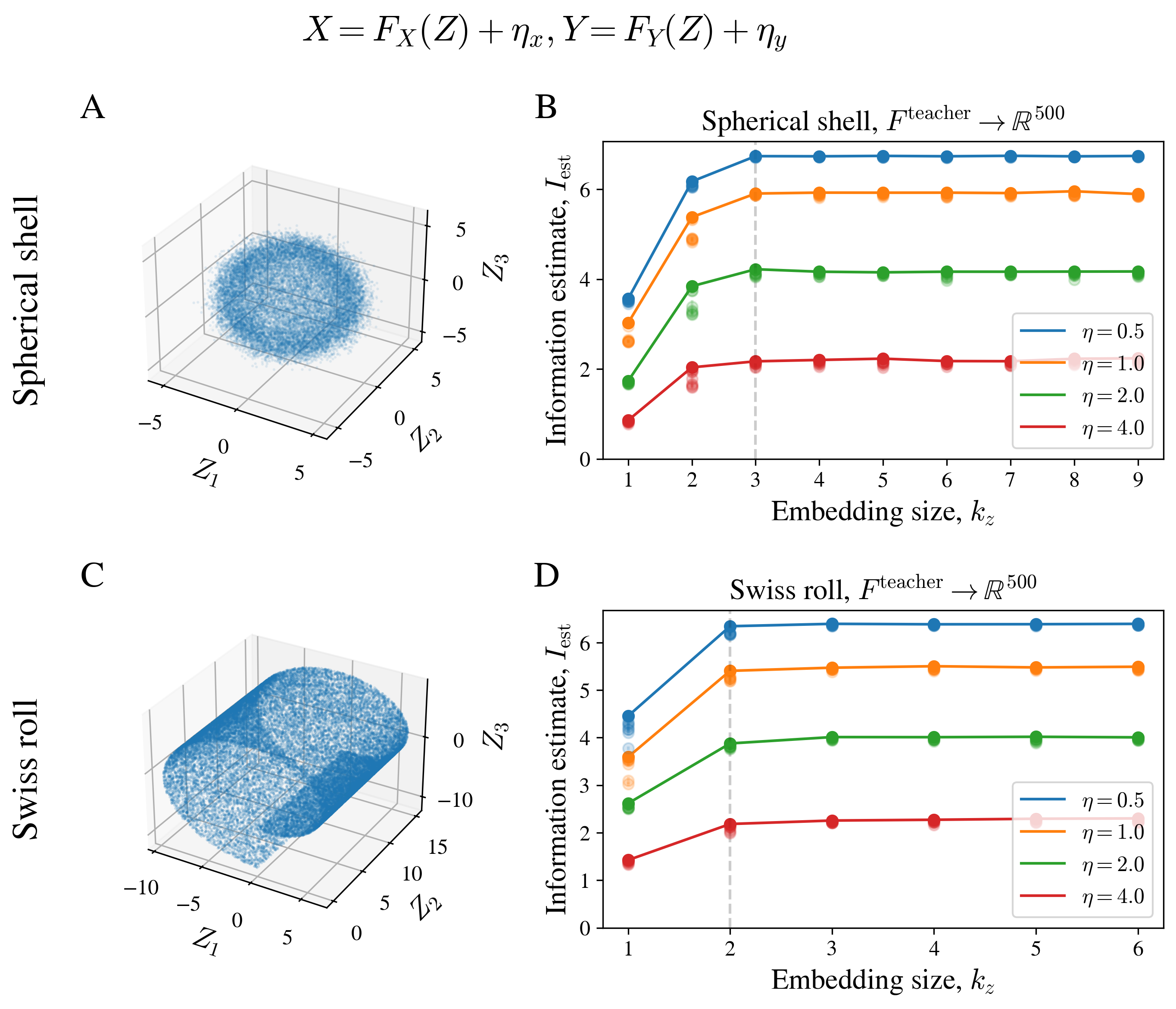}
    \caption{\textbf{Estimating intrinsic dimensionality by view splitting.} We infer the
    dimensionality of a shared latent variable $Z$ from two noisy views $X$ and $Y$.
    \textbf{(A,C)} Latent manifolds: (A) hypersphere ($K_Z=3$) and (C) Swiss roll ($K_Z=2$).
    \textbf{(B,D)} Estimated MI versus $k_z$. With observation noise $\eta>0$, the estimate saturates
    at the true latent dimensionality $K_Z$, recovering the degrees of freedom of the shared signal. Error bars are standard deviations over 10 trials (semi-transparent markers).}
    \label{fig:shared_latent}
\end{figure}

We validate this on two challenging latent topologies: a hypersphere ($K_Z=3$) and a Swiss roll
($K_Z=2$). Two high-dimensional noisy views are generated as in Eq.~(\ref{eq:noise_view_splitting})
(see App.~\ref{app:details_synthetic_data} for details). As shown in Fig.~\ref{fig:shared_latent},
even when the observation maps project these manifolds into $K=500$ dimensions, our estimator
identifies the saturation point at $k_z=K_Z$.

This justifies the view-splitting strategy employed in Sec.~\ref{subsec:results-real}. While for the
pendulum, taking past as $X$ and future as $Y$ is natural, for the Ising model, we can instead partition
each configuration into two spatial domains, yielding views that are conditionally independent given
the latent state $Z$. Our estimator then recovers the dimensionality of this shared latent structure.

\subsection{Comparing with intrinsic dimensionality estimators}
\label{app:compare_with_ID}

\begin{figure}[h]
    \centering
    \includegraphics[width=0.55\linewidth]{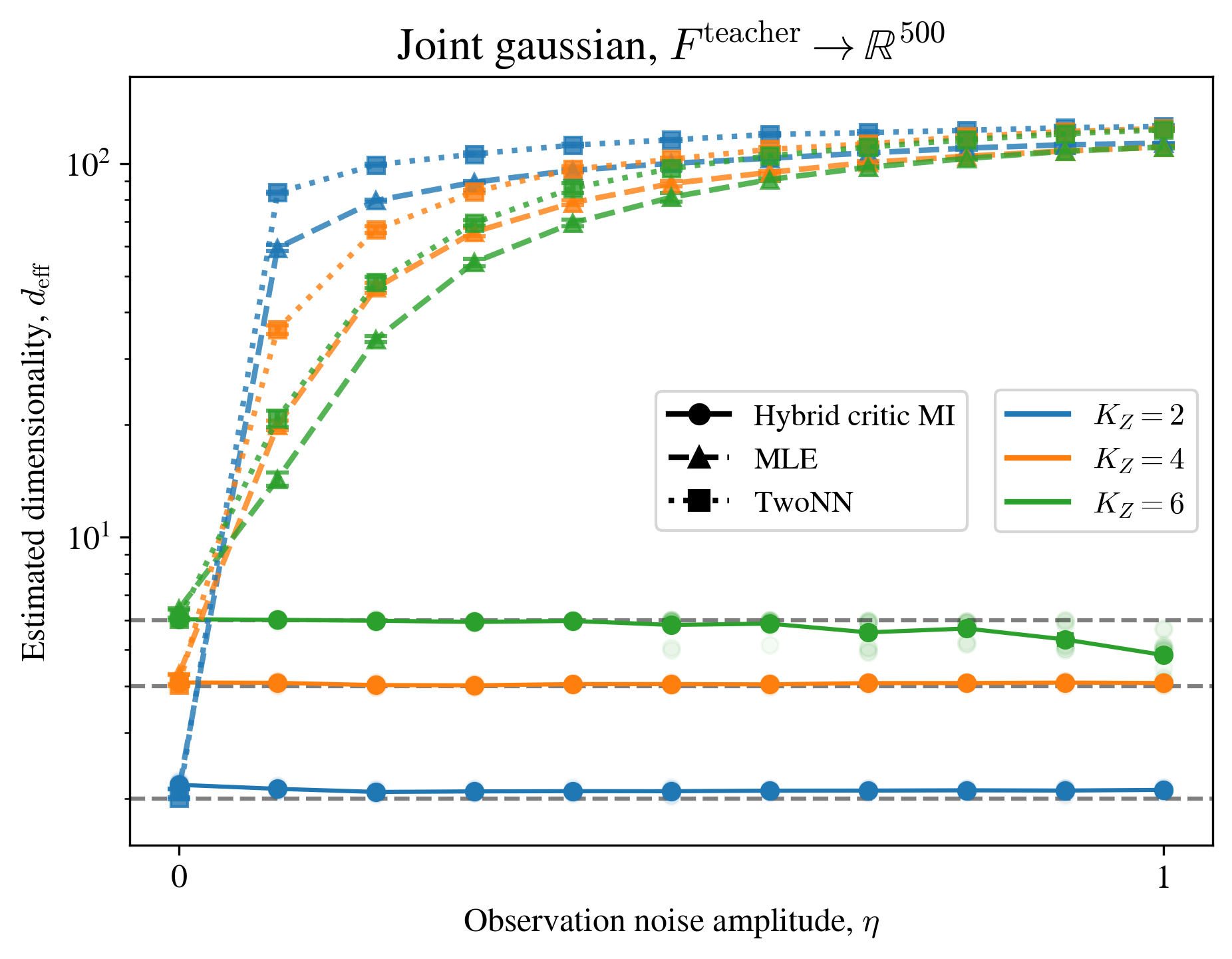}
    \caption{\textbf{Comparison of intrinsic dimensionality estimators with our paired-view MI protocol.}
We use a finite dataset of size $N=16{,}384$ generated from a $K_Z$-dimensional Gaussian latent,
mapped through teacher transforms into $K=500$ observed dimensions, with added uncorrelated
observation noise. In the absence of noise, all methods recover the latent dimension. With
observation noise ($\eta>0$), MLE and Two-NN degrade and cannot decouple latent dimensionality from
observation-space noise, while our hybrid-critic MI protocol remains robust over the range shown.
For MLE and Two-NN, we report the average of the estimates obtained from applying the estimator to
$X$ and to $Y$. All methods averaging over 10 independent trials; error bars are standard deviations.}
    \label{fig:compare_with_id}
\end{figure}

We compare our view-splitting, hybrid-critic MI dimensionality protocol with standard intrinsic
dimensionality (ID) estimators: the Levina--Bickel maximum-likelihood estimator (MLE)
\citep{levina2004maximum} (with an adaptive neighborhood size) and the Two--Nearest-Neighbor
estimator (Two-NN) \citep{facco2017estimating}, implemented using the \texttt{scikit-dimension}
library \citep{bac2021scikit}. In Fig.~\ref{fig:compare_with_id} we use the same synthetic
construction as in Sec.~\ref{subsec:infinite-noise}: data are generated from a jointly Gaussian
latent, passed through the teacher network, and corrupted by additive observation noise. The ID
estimators are applied separately to the $X$ and $Y$ point clouds, whereas our task-relevant
dimensionality estimate is computed from the paired dataset $(X,Y)$.

In this setting, all estimators recover the latent dimensionality in the absence of added
observation noise ($\eta=0$). With observation noise, however, the geometric ID estimators degrade
sharply, tending towards the ambient observation dimension. In
contrast, our protocol---computing $d_{\rm eff}$ via the participation ratio of the learned latent
cross-covariance spectrum---remains robust and continues to recover the latent dimensionality over
the noise levels tested.

While this behavior is expected (ID estimators cannot  separate latent structure from
observation noise, whereas the paired-view formulation targets shared signal), it highlights the
practical value of \emph{task-relevant} dimensionality: for high-dimensional scientific data mired
in observational noise, the paired-view MI approach can remain informative in regimes where
standard ID estimators do not.

\section{Additional Results}

\subsection{Non-invertibility of the Gaussian mixture}
\label{app:non_invert_mix}

\begin{figure}[t]
    \centering
    \includegraphics[width=0.45\linewidth]{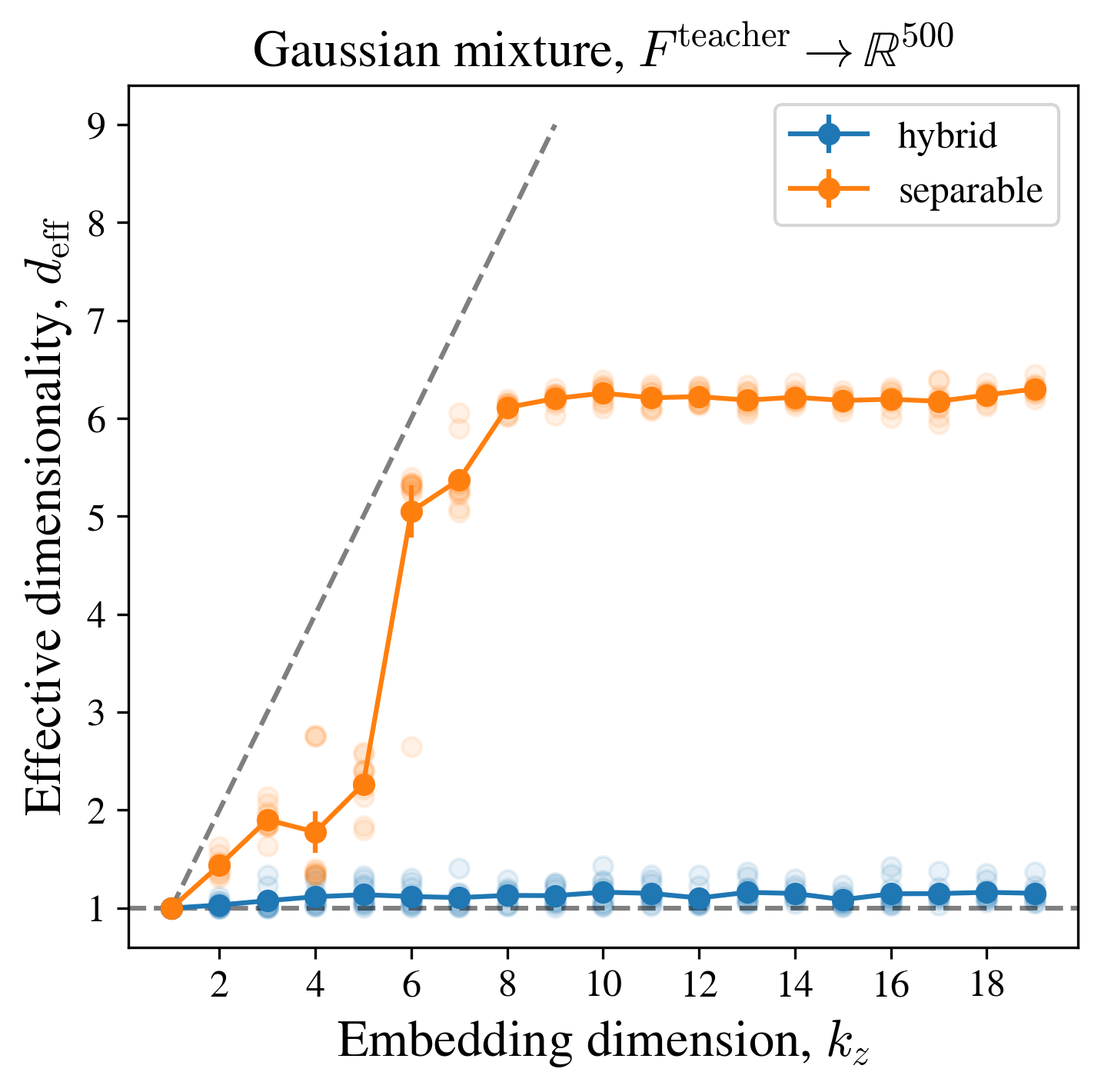}
    \caption{\textbf{Separable critics inflate dimensionality for multimodal latents.}
    For the representative Gaussian mixture ($N_{\rm peaks}=8$, $\mu=2.0$, $I_p=2.0$ bits), the
    separable critic yields an inflated effective dimensionality, $d_{\rm eff}\approx 7$,
    whereas the hybrid critic recovers the latent dimensionality. We conjecture that the inflation
    reflects the need to represent a multimodal dependence structure using bilinear product terms.  Averages over 10 independent trials; error bars are standard deviations.}
    \label{fig:si_deff_separable_gauss_mix_representative}
\end{figure}

The Gaussian mixture distribution (see App.~\ref{app:latent_dist_specifications} for the density),
constructed to have multiple overlapping clusters, is characterized by the number of clusters
$N_{\rm peaks}$, the ring radius $\mu$ in $Z_X$--$Z_Y$ space, and the within-cluster correlation
$\rho$. For $N_{\rm peaks}>1$ the joint distribution is multimodal, and the conditional structure is
generically non-invertible. As shown in the main text (Fig.~\ref{fig:fig2_infinite_synthetic_joint}E,F)
for our representative mixture ($N_{\rm peaks}=8$, $\mu=2.0$, $\rho\approx 0.97$), dimensionality
estimates based on MI saturation with a separable critic are inflated. We see now that the same conclusion holds for
the participation-ratio method developed in this paper (Fig.~\ref{fig:si_deff_separable_gauss_mix_representative}).

\begin{figure}[t]
    \centering
    \includegraphics[width=0.7\linewidth]{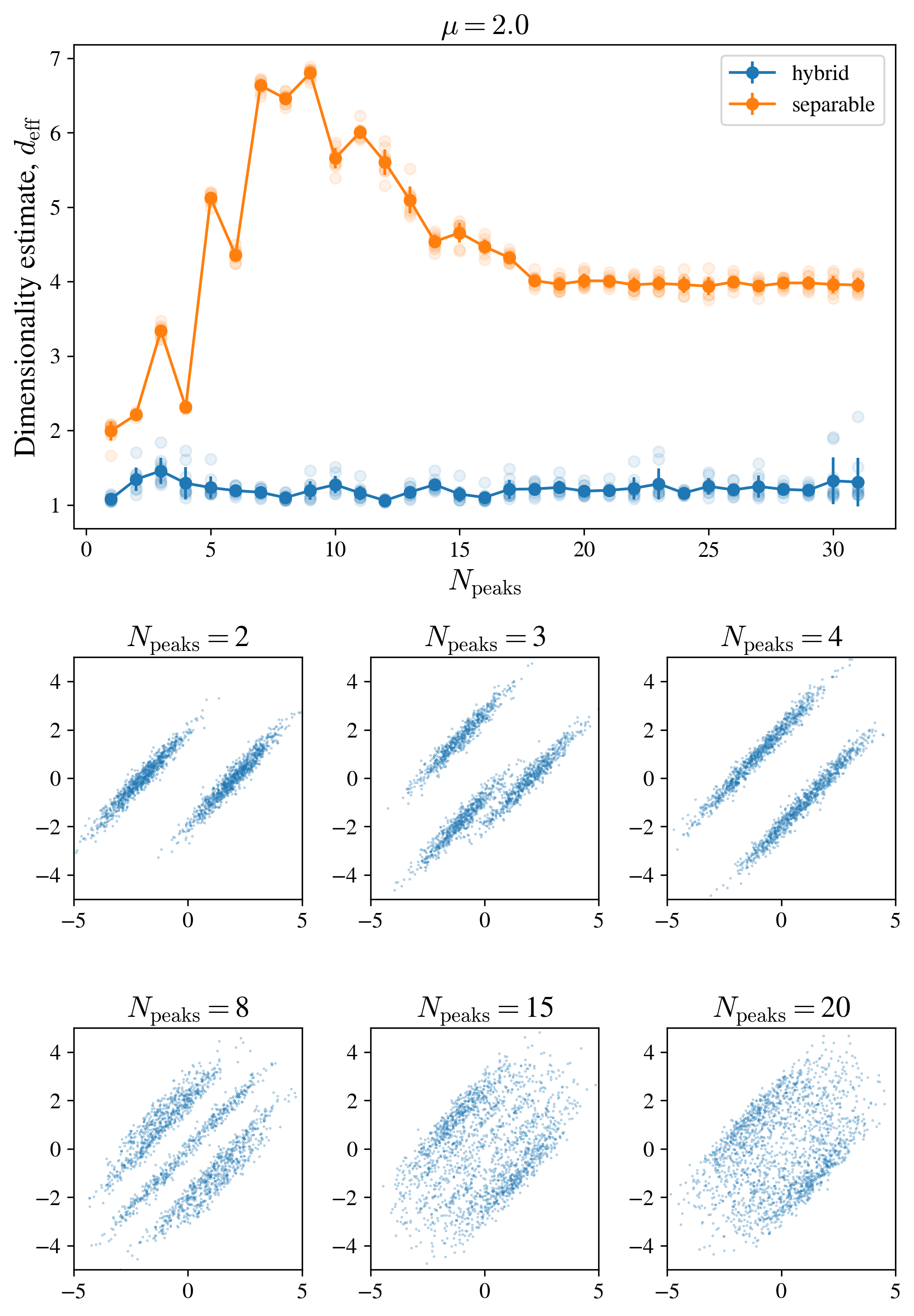}
    \caption{\textbf{Effective dimensionality of the Gaussian mixture versus number of clusters.}
    \textbf{(top)} Across mixtures with increasing $N_{\rm peaks}$, the hybrid critic ($k_z=64$) tracks the
    latent dimensionality ($K_Z=1$), while the separable estimate shows a non-monotonic dependence on
    $N_{\rm peaks}$. Under the patchwise interpretation discussed in the text, the separable
    dimensionality reflects the number of approximately Gaussian regions needed to represent the
    multimodal dependence. This is consistent with the drop in the separable estimate from
    $N_{\rm peaks}=3$ to $4$ (see representative samples, \textbf{(bottom)}) and from $5$ to $6$ (not shown). As always, semi-transparent markers denote individual trials.}
\label{fig:d_eff_separable_invertibility_peaks}
\end{figure}

We further conjecture that this inflation reflects an implicit  decomposition of the latent
space into patched  within which the joint distribution is approximately Gaussian, so that a
bilinear critic can represent the dependence locally. In this view, the minimal number of such
regions (e.g., admitting approximate Gaussian copulas) sets the effective dimensionality required by
the separable critic \citep{kulpa1999approximation,ghosh2009patchwork}. This interpretation is
consistent with the dependence of the separable $d_{\rm eff}$ on the number of mixture components
(Fig.~\ref{fig:d_eff_separable_invertibility_peaks}). For example, for our parameters, the estimated
$d_{\rm eff}$ decreases at $N_{\rm peaks}=4$ as clusters collapse pairwise due to the ring symmetry
(see representative samples in Fig.~\ref{fig:d_eff_separable_invertibility_peaks}), and similarly at
$N_{\rm peaks}=6$ (not shown). Under this interpretation, separable critics may also provide an empirical
probe of multimodality in low-dimensional latent distributions; we defer a quantitative analysis to
future work.

% \FloatBarrier

\subsection{Smaller embedding space: fewer samples to estimate MI}

\begin{figure}[t]
    \centering
    \includegraphics[width=0.45\linewidth]{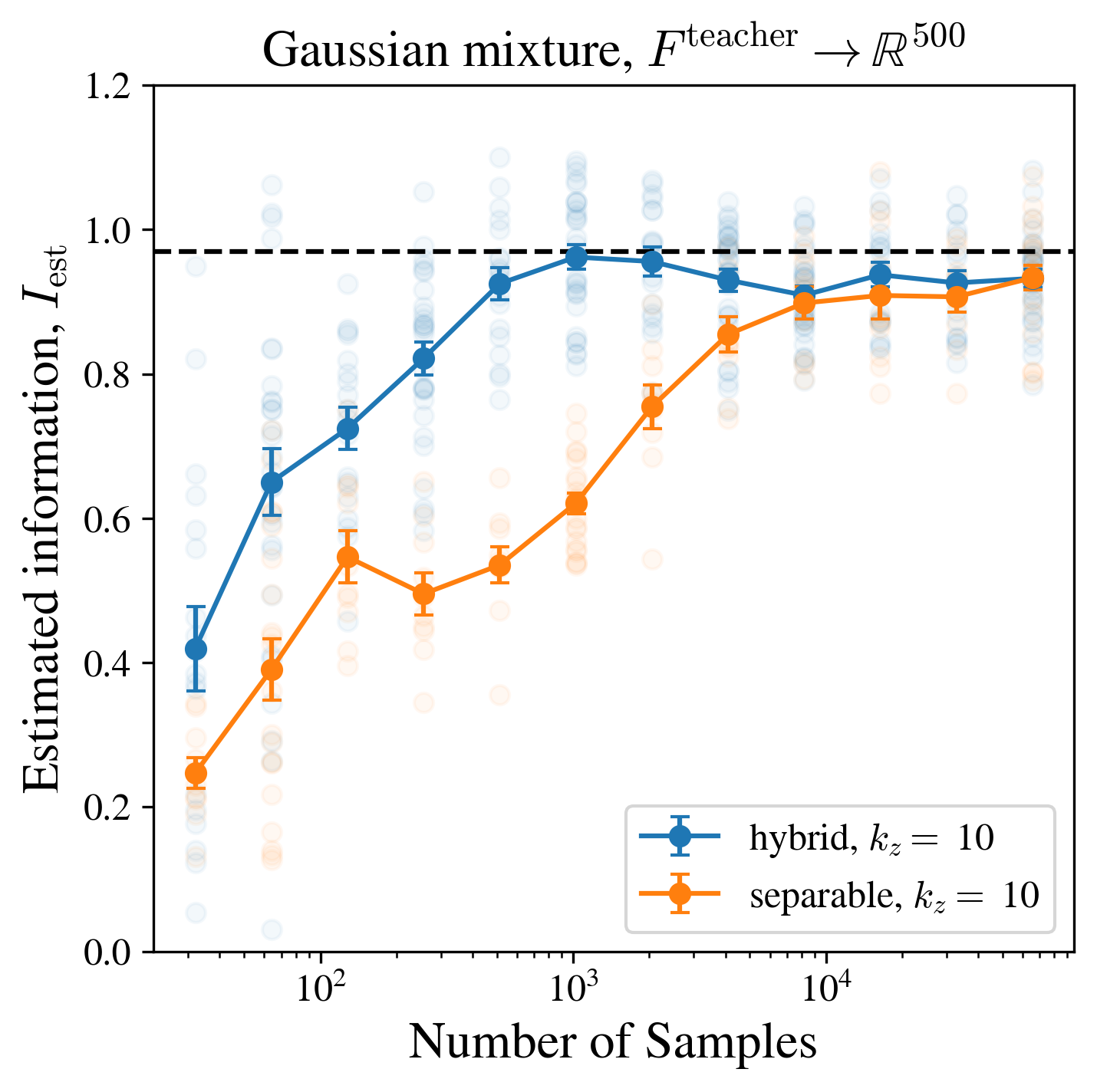}
   \caption{\textbf{Sample efficiency for MI estimation on the Gaussian mixture.}
For the representative Gaussian mixture, the hybrid critic learns a lower-dimensional embedding and
consequently reaches accurate MI estimates with fewer samples than a separable critic. Markers show
individual trials (semi-transparent), and curves denote the mean / standard deviations  across trials.}
    \label{fig:si_gauss_mix_MI_samples}
\end{figure}

Using a hybrid critic can be beneficial even when the goal is not dimensionality estimation. For
multimodal latent distributions such as our representative Gaussian mixture, a separable critic
learns an effectively higher-dimensional embedding ($d_{\rm eff}\approx 7$) than the hybrid critic
($d_{\rm eff}\approx 1$). This increase in effective dimension directly translates into a higher
sample requirement for accurate MI estimation: higher-dimensional embeddings are harder to populate,
and MI estimation error grows with the effective dimensionality of the dependence structure, as
argued in \citet{abdelaleem2025accurate} and in related sample-efficiency analyses of simultaneous reduction methods \citep{abdelaleem2024simultaneous} and information bottleneck objectives \citep{martini2024data}. Indeed,
Fig.~\ref{fig:si_gauss_mix_MI_samples} shows that the hybrid critic reaches accurate MI estimates
with substantially fewer samples than the separable critic for the same underlying mixture
distribution. We therefore conclude that the hybrid architecture can improve not only
dimensionality identifiability but also practical MI estimation from limited data by reducing the
effective embedding dimension.

\subsection{2D Ising model}

\begin{figure}[t]
    \centering
    \includegraphics[width=.8\linewidth]{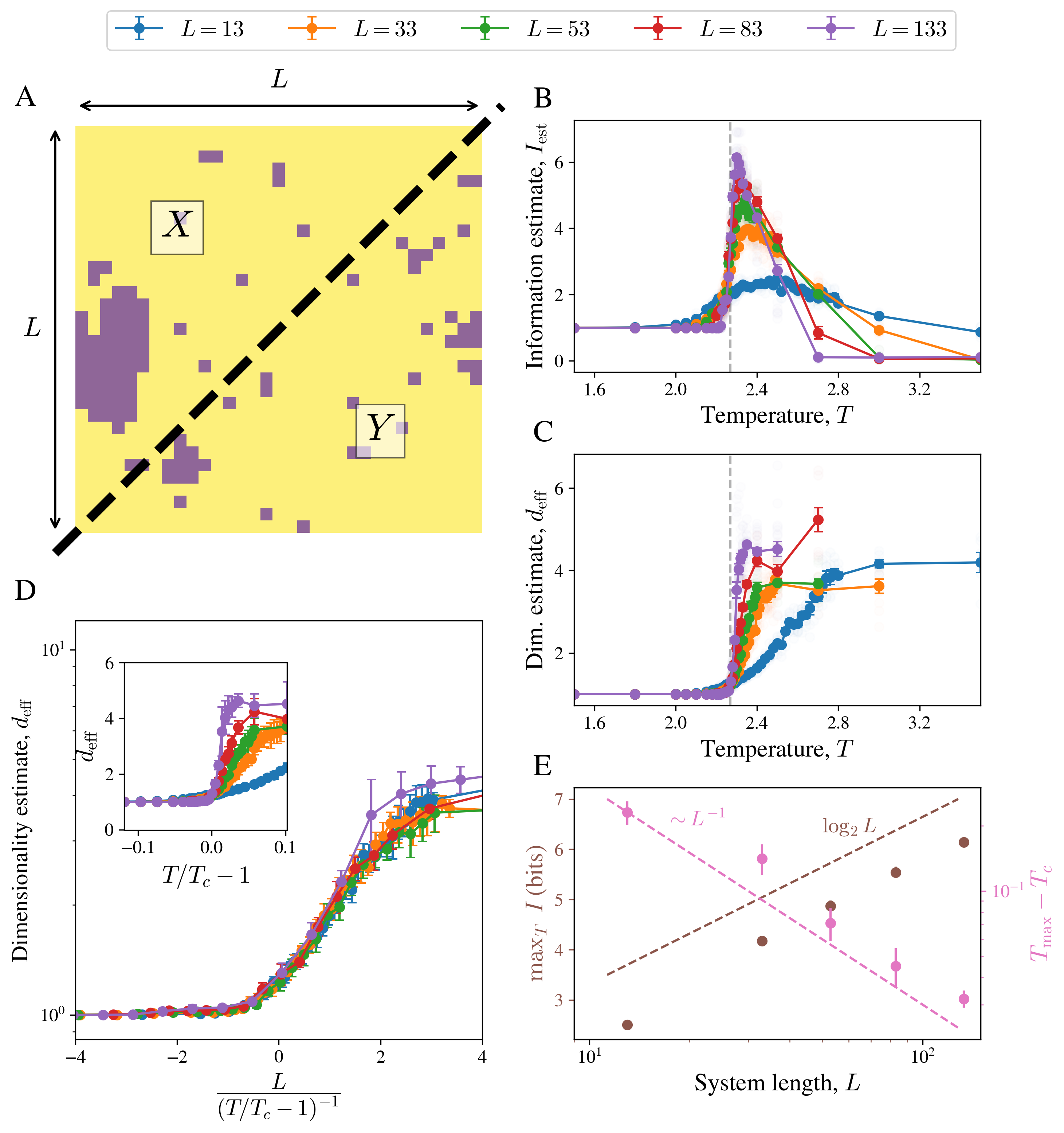}
    \caption{\textbf{Ising model: mutual information and dimensionality.}
    \textbf{(A)} Schematic illustrating a spin configuration and the spatial partitioning used to
    construct paired datasets. At each temperature $T$ and system size $L$, we generate $10{,}000$
    approximately independent equilibrium configurations of the 2D nearest-neighbor Ising model
    ($J=1.0$, $h=0.0$) using Markov chain Monte Carlo (see App.~\ref{app:details_ising}).
    \textbf{(B)} The estimated MI using our MI estimator (with $k_z=64$) peaks near the known
    critical temperature $T_c\approx 2.269$.
    \textbf{(C)} The corresponding effective dimensionality shared between the two spatial
    partitions. Below $T_c$ it approaches $\sim 1$, reflecting the macroscopic magnetization degree
    of freedom in the ordered phase, and increases as correlated domains grow near criticality.
    \textbf{(D)} Near criticality the correlation length diverges as
    $\xi\sim |T-T_c|^{-\nu}$ with $\nu=1$, so $d_{\rm eff}\sim L/\xi$ implies the collapse of
    $d_{\rm eff}$ across $L$ when plotted against the standard scaling variable
    $L^{1/\nu}(T/T_c-1)$.
    \textbf{(E)} From (B) we extract $I_{\max}$ and $T_{\max}$ for each $L$, with means and standard
    deviations obtained by bootstrapping across trials. This recovers the leading-order finite-size
    scalings $I_{\max}\sim \log L=\tfrac{1}{2}\log N$ (with $N=L^2$ and prefactor $\tfrac{1}{2}$
    corresponding to one macroscopic degree of freedom, the magnetization
    \citep{tchernookov2013predictive}) and $T_{\max}-T_c\sim L^{-\nu}$ with $\nu=1$. Throughout, solid
    markers denote means over 10 independent training trials with error bars showing the standard
    deviation; in (B,C) semi-transparent markers show individual trials.}
    \label{fig:ising_supplementary}
\end{figure}

Additional results from training on the simulated Ising-model spin configurations are shown here,
including the MI as a function of temperature (Fig.~\ref{fig:ising_supplementary}). For $T<T_c$
(in the ferromagnetically ordered phase at $h=0$), the MI estimate approaches $\sim 1$ bit,
consistent with the two symmetry-related magnetization sectors, and it drops toward zero for
$T>T_c$ in the disordered phase (Fig.~\ref{fig:ising_supplementary}B). We also show the unscaled
dimensionality estimate $d_{\rm eff}$ for different system sizes (Fig.~\ref{fig:ising_supplementary}C),
displaying $d_{\rm eff}$ only when the MI estimate is above a reliability threshold
(here $I_{\rm est}>0.5$ bits). The peak MI, $I_{\max}=\max_T I$, and the peak location $T_{\max}$
follow the expected finite-size scaling behavior, $I_{\max}\sim \log L \sim \log N$ and
$T_{\max}-T_c\sim L^{-\nu}=L^{-1}$, as shown in Fig.~\ref{fig:ising_supplementary}E
\citep{goldenfeld2018lectures,bialek2001predictability,tchernookov2013predictive}.

%\FloatBarrier
% ==============================================================================
% APPENDIX D: EXPERIMENTAL DETAILS (SYNTHETIC)
% ==============================================================================
\section{Experimental Details: Synthetic Data}
\label{app:details_synthetic}
Here we provide the specific hyperparameters, architectures, and training protocols used for the synthetic benchmarks in Section~\ref{sec:results-synthetic}.

\subsection{Data generation and transformations}
\label{app:details_synthetic_data}

To simulate high-dimensional observations, a
low-dimensional latent variable $Z\in\mathbb{R}^{K_Z}$ is drawn from a ground-truth distribution and
mapped to high-dimensional observation vectors $X,Y\in\mathbb{R}^{500}$ via frozen, randomly
initialized neural networks (``Teacher'' networks).

\paragraph{Latent Distributions.}
\label{app:latent_dist_specifications}
In the text and the appendices, we discuss the following joint or shared latent distributions:
\begin{enumerate}
    \item {Joint Gaussian:} A standard baseline with total MI $I$ (in bits) and latent dimensionality
    $K_Z$, unit variance and zero mean. This MI is split equally across dimensions, i.e.
    $\rho = \sqrt{1-2^{-2I/K_Z}}$.
    \begin{equation}
    p(z_x,z_y)
    =
    \mathcal N\!\left(
    \begin{bmatrix} z_x \\ z_y \end{bmatrix};
    0,
    \begin{pmatrix}
    \mathbb I_{K_Z} & \rho\,\mathbb I_{K_Z} \\
    \rho\,\mathbb I_{K_Z} & \mathbb I_{K_Z}
    \end{pmatrix}
    \right).
    \end{equation}
    Unless otherwise specified, we use $I=2.0$ bits and $K_Z=4$.
    
    \item {Gaussian Mixture:} A complex, multi-modal distribution deliberately designed to be challenging
    with $K_Z=1$. It has $N_{\rm peaks}$ equally likely, unit-variance jointly Gaussian components, with
    means equally spaced on a ring of radius $\mu$ in $(z_x,z_y)$ space, and per-component correlation $\rho_{\rm peak}$:
    \begin{equation}
    p(z_x,z_y)
    =
    \frac{1}{N_{\rm peaks}}
    \sum_{k=1}^{N_{\rm peaks}}
    \mathcal N\!\left(
    \begin{bmatrix} z_x \\ z_y \end{bmatrix};
    \begin{bmatrix} \mu\cos\theta_k \\ \mu\sin\theta_k \end{bmatrix},
    \begin{pmatrix}
    1 & \rho_{\rm peak} \\
    \rho_{\rm peak} & 1
    \end{pmatrix}
    \right),
    \quad
    \theta_k=\frac{2\pi k}{N_{\rm peaks}},
    \end{equation}
    where $\rho_{\rm peak} = \sqrt{1-2^{-2I_{\rm peak}}}$.
    Unless otherwise specified, $N_{\rm peaks} = 8$, $\mu=2.0$, and $I_{\rm peak} = 2.0$ bits, equivalent to $\rho_{\rm peak}\approx 0.97$.
    
    \item {Noisy Hyperspherical Shell:}
    This is used in the shared-latent setting, with $z$ sampled near a hyperspherical surface
    $S^{K_Z-1}\subset\mathbb{R}^{K_Z}$ of radius $r$, with radial noise of variance $\sigma_r^2$:
    \begin{equation}
    z = (r+\epsilon_r)\,\hat n,
    \qquad
    \hat n \sim \mathrm{Unif}(S^{K_Z-1}),\quad
    \epsilon_r\sim\mathcal N(0,\sigma_r^2).
    \end{equation}
    Unless otherwise specified, $K_Z=3$, $r=4$, $\sigma_r=0.5$.
    
    \item {Swiss Roll:} A standard curved manifold of intrinsic dimension 2 embedded in $\mathbb{R}^3$,
    generated by $t\sim \mathrm{Unif}(t_0,t_1)$ and $h\sim \mathrm{Unif}(h_0,h_1)$:
    \begin{equation}
        Z_1 = t\sin t, \qquad Z_2 = t\cos t, \qquad Z_3 = h\;.
    \end{equation}
    Unless otherwise specified, $t_0 = 1.5 \pi$, $t_1 = 3.5 \pi$, $h_0 =0$, $h_1 =15$.
\end{enumerate}

\paragraph{The Teacher Networks.}
The mapping functions $F_X, F_Y: \mathbb{R}^{K_Z} \to \mathbb{R}^{500}$ are parameterized as
Multi-Layer Perceptrons (MLPs). Unless otherwise stated (e.g., the ``Linear'' baseline in
Fig.~\ref{fig:fig2_infinite_synthetic_joint}B,E), these networks have one hidden layer of size 1024,
with Xavier normal initialization and Softplus activation to induce nonlinear structure in the
observation space. We add white noise relative to the signal strength where specified (e.g., in
Fig.~\ref{fig:fig3_infinte_synthetic_joint_noisy}).

\subsection{Critic architectures}
\label{app:details_synthetic_arch}

All critic architectures use LeakyReLU activations in all layers except the final output layer.
All layers are initialized with Xavier uniform initialization.

\paragraph{Separable Critic.}
The separable architecture, defined as $T(x,y)=g^X(x)^\top g^Y(y)$, uses independent encoders for
$X$ and $Y$. These encoders are parameterized as two-layer MLPs with 128 hidden units, mapping the
input to an embedding of dimension $k_z$, which is swept (or fixed to a large value) across
experiments.

\paragraph{Hybrid Critic.}
Our hybrid architecture, $T(x,y)=T_\theta([g^X(x),g^Y(y)])$, retains the same encoder backbone as
the separable critic but fuses the embeddings via a concatenated head. This mixing head $T_\theta$
is an MLP that takes the concatenated embeddings $[g^X(x),g^Y(y)]$ as input and processes them
through a single hidden layer with 64 units.

\subsection{Training protocols}
\label{app:details_synthetic_training}

All networks are implemented in \texttt{PyTorch} and optimized using Adam , a learning rate
$5\times 10^{-4}$.

\paragraph{Infinite Data Regime.}
This is the regime used for Figs.~\ref{fig:fig2_infinite_synthetic_joint},
\ref{fig:fig3_infinte_synthetic_joint_noisy}, \ref{fig:shared_latent}, and
\ref{fig:fig5_pr_infinite_synthetic}. Because a fresh batch is generated at every training step,
there is no overfitting. Estimators are trained for 20{,}000 iterations with batch size 128. The
reported MI is the average over the final 10\% of training steps.

\paragraph{Finite Data Regime.}
For the finite-data experiments in Fig.~\ref{fig:fig6_finite_synthetic}, we train on fixed datasets
with sample size $N$ as indicated in the figure. We train for 100 epochs, evaluating a held-out
test set of 128 samples at each epoch. We then use the Max-Test heuristic
(App.~\ref{app:protocol_maxtest}), selecting the checkpoint that maximizes the MI estimate on this
test set.

\paragraph{Computational Resources.}
Experiments were run on AWS instances of type L4, L40s, A100, and H200, using configurations
appropriate to the scale of each sweep. Runtime varies with dataset size and training parameters.
For example, one trial in the infinite-data resampling regime (Fig.~\ref{fig:fig2_infinite_synthetic_joint})
with $K=500$ observed dimensions and $20{,}000$ iterations at batch size 128 takes $\sim 70$ seconds,
while a single-pendulum run (Fig.~\ref{fig:real_data}) for one trial of 200 epochs with 1{,}000
initial conditions (57{,}000 samples of size $2\times 128\times 128$) takes $\sim 3{,}000$ seconds.

% ==============================================================================
% APPENDIX E: EXPERIMENTAL DETAILS (PHYSICAL SYSTEMS)
% ==============================================================================
\section{Experimental Details: Physical Systems}
\label{app:details_physics}

\subsection{2D Ising model}
\label{app:details_ising}
We generate spin configurations using Markov Chain Monte Carlo (MCMC) simulations of the two–dimensional ferromagnetic Ising model with local single–spin Metropolis updates~\citep{metropolis1953equation, newman1999monte} on a
square lattice of linear size $L$ with periodic boundary conditions. For each temperature $T$,
configurations are sampled using single-spin stochastic updates that satisfy detailed balance,
producing equilibrium ensembles of spin states. One Monte Carlo sweep consists of sequential updates
on the two checkerboard sublattices. Multiple independent replicas are evolved in parallel by
embedding several $L\times L$ systems into a larger lattice; each replica is initialized in a
uniform up or down state chosen at random and evolved independently. After a fixed number of sweeps
between measurements, observables are recorded separately for each replica, yielding ensembles of
approximately decorrelated configurations.

Throughout this work we restrict to zero external field ($h=0$) and unit coupling ($J=1$). We sweep
system sizes from $L=13$ to $L=133$ and temperatures from $T=0$ to $T=4.5$, spanning both ordered and
disordered phases and crossing the critical region. For each $(L,T)$, we generate collections of
$10{,}000$ equilibrium configurations by running fixed blocks of Monte Carlo sweeps between
measurements, and treat each replica as an independent sample. These spin configurations form the
datasets used in the main text.

At each $(L,T)$, we construct paired views $(X,Y)$ by spatially splitting the lattice as shown in
the main text, and train the MI estimator. The encoders and concatenated head use the same
architecture as in the synthetic benchmarks (encoders: 2-layer MLP with 128 hidden units;
concatenated head: 1-hidden-layer MLP with 64 units; LeakyReLU activations; Xavier initialization).
Models are trained with batch size 128 for up to 100 epochs, using the max-test stopping criterion.
We use 10\% of samples as a test set and report $\hat I_{\text{train}}(t^\ast)$ at the selected
epoch. Dimensionality is reported as the participation ratio of the singular values of the
cross-covariance of the trained encoder representations at the max-test epoch.

\subsection{Pendulum dynamics}
\label{app:details_pendulum}

To evaluate our estimator on high-dimensional data with known physical degrees of freedom, we use
the video dataset of \citet{chen2022automated}, focusing on the Single Pendulum and the Rigid Double
Pendulum. The basic setup---constructing paired views by predicting future from past in a dynamical
system---has close analogues in dynamical-systems theory (e.g., delay embeddings and data-driven
dynamics) \citep{takens2006detecting,schmid2022dynamic}, in predictive-information formulations
\citep{bialek2001predictability,meng2022compressed}, and in modern representation-learning approaches based on
predictive objectives \citep{oord2018representation,mardt2018vampnets,lecun2022path}.

\paragraph{Physical Parameters.}
The Single Pendulum has mass $m=1$\,kg and length $L=0.5$\,m; its state is $(\theta,\dot\theta)$,
so that true $d=2$. The Rigid Double Pendulum consists of two arms ($L_1=20.5$\,cm, $L_2=17.9$\,cm,
$m_1=0.262$\,kg, $m_2=0.11$\,kg) and has state $(\theta_1,\theta_2,\dot\theta_1,\dot\theta_2)$, so that true
$d=4$.

\paragraph{Data Preprocessing.}
Videos are $128\times128$ RGB frames, denoted as $\Phi$. We convert to grayscale, so that $\Phi \in \mathbb{R}^{128\times128} $. To form paired views, we use time:
since velocities are not determined by a single frame, we perform delayed embedding \citep{takens2006detecting} by concatenating consecutive frames and defining
$X=[\Phi_t,\Phi_{t+1}]$ and $Y=[\Phi_{t+2},\Phi_{t+3}]$.

\paragraph{Dataset Organization.}
Data are organized into folders, each containing 60 consecutive frames from one trajectory (one
initial condition). We use 1,100 trajectories total, split into 1,000 training folders and 100 test
folders. The pairing above yields 57 samples per trajectory, giving $N_{\text{train}}=57{,}000$
pairs and $N_{\text{test}}=5{,}700$ pairs.

\paragraph{Model Architecture.}
Because $X$ and $Y$ are the same system at nearby times, we enforce a Siamese constraint $g^X=g^Y$.
We use the same backbone as elsewhere: an MLP encoder mapping the $2\times128\times128$ input to an
embedding of dimension $k_z=64$, and the critic head as in App.~\ref{app:details_synthetic_arch},
with the addition of LayerNorm in the critic to stabilize training.

\begin{figure}[t]
    \centering
    \includegraphics[width=\linewidth]{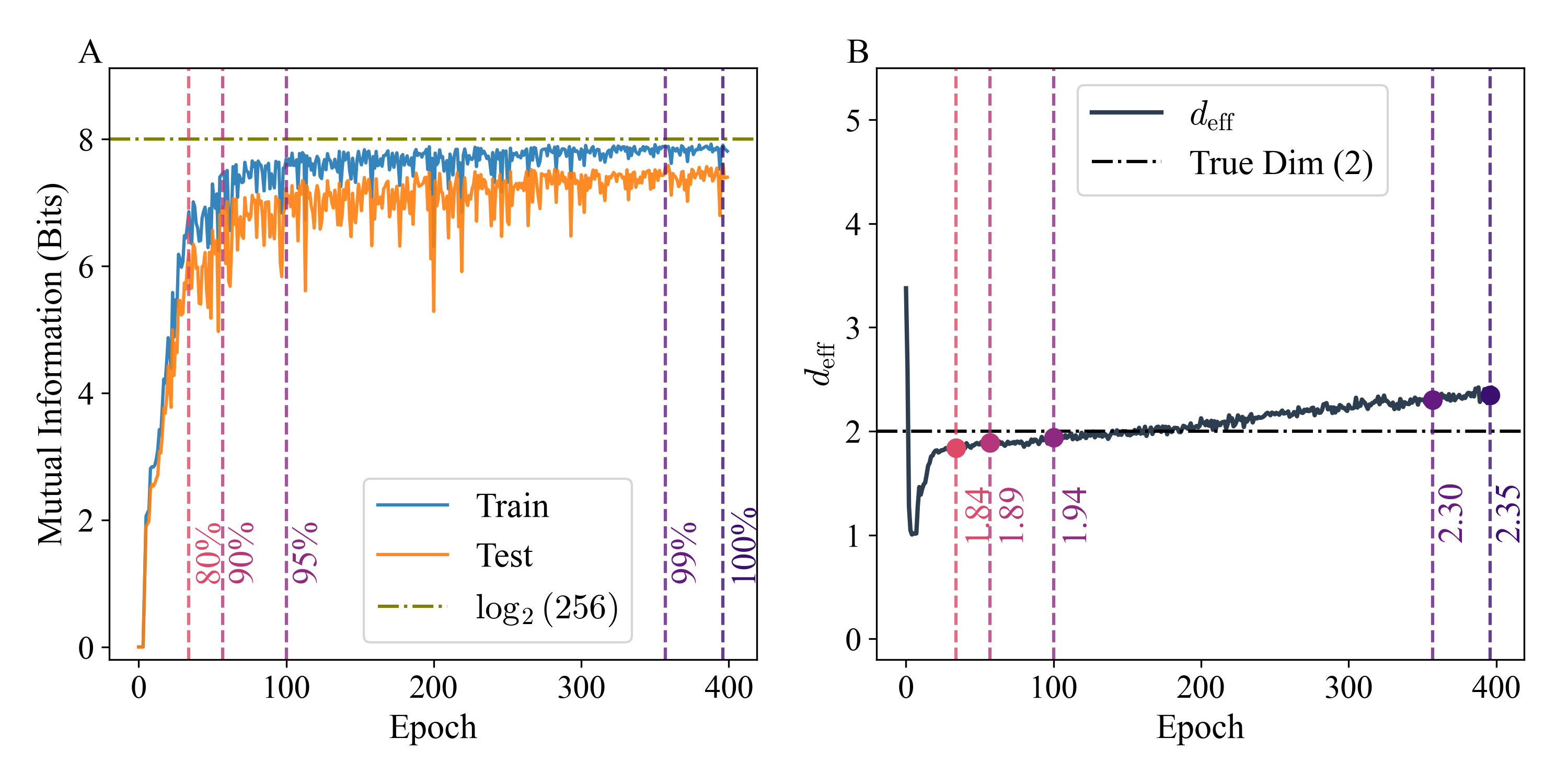}
    \caption{\textbf{Saturation and stopping in the pendulum setting.}
    Single-pendulum example (trained on 500 trajectories).
    \textbf{(A)} Train/test MI versus epoch; both approach the InfoNCE ceiling
    $\log_2(\text{batch size})\approx 8$ bits. Vertical lines denote stopping points at different
    fractions of the maximum test MI.
    \textbf{(B)} $d_{\rm eff}$ versus epoch. Although prolonged training can gradually inflate
    $d_{\rm eff}$, the estimate remains near the true value ($d_{\rm eff}\approx 2$) across a wide
    range of stopping thresholds.}
    \label{fig:supp_pendulum}
\end{figure}

\begin{figure}[t]
    \centering
    \includegraphics[width=\linewidth]{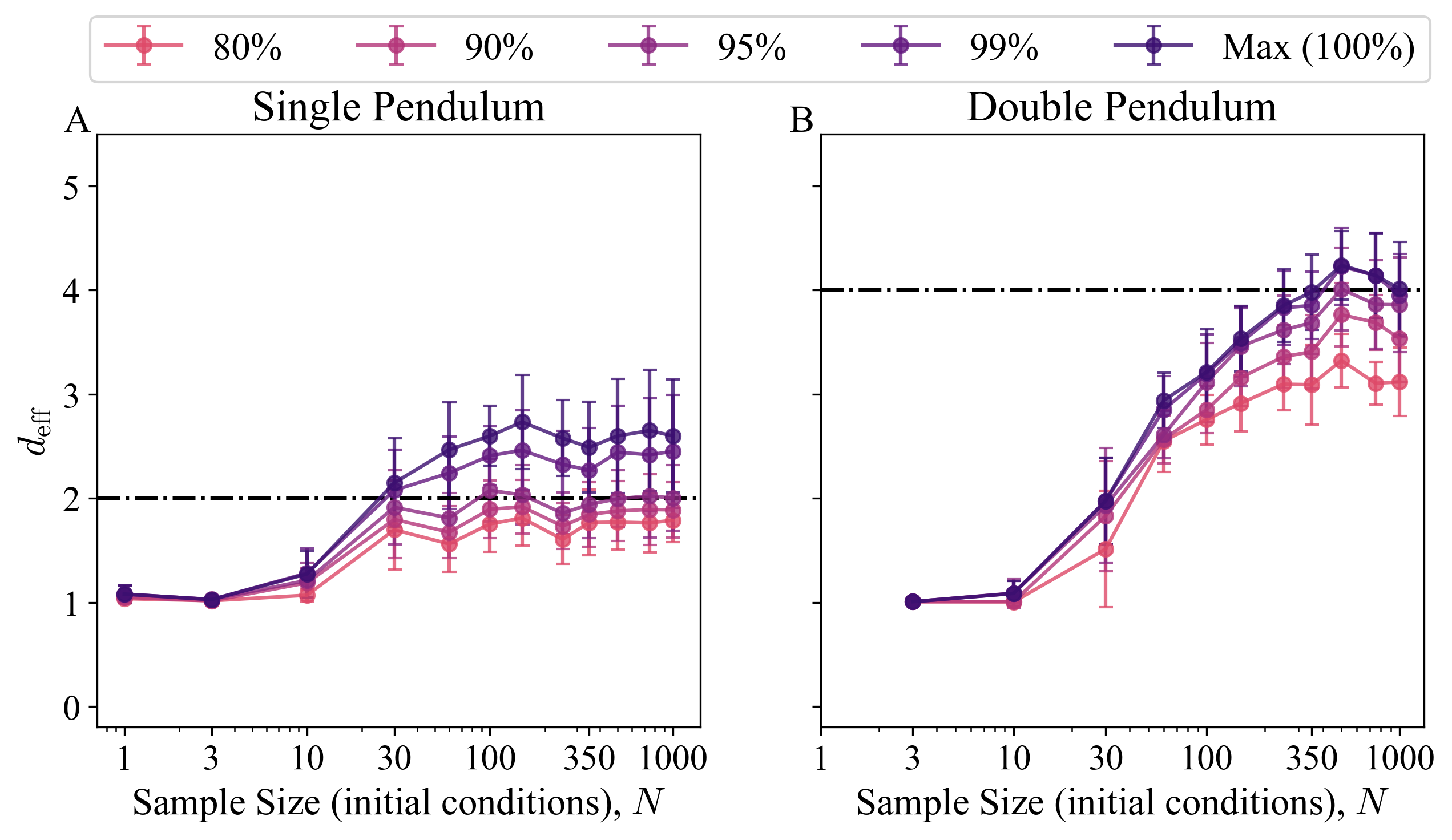}
    \caption{\textbf{Robustness to stopping thresholds.}
    Extension of Fig.~\ref{fig:real_data}B across dataset sizes (up to 1,000 trajectories) and
    stopping thresholds (fractions of max test MI, $80\%$--$100\%$). Estimated dimensionalities are
    stable: $d_{\rm eff}\approx 2$ for the single pendulum (inflating toward 3 at very late
    thresholds) and $d_{\rm eff}\approx 4$ for the double pendulum near the maximum threshold.}
    \label{fig:supp_pendulum_threshs}
\end{figure}

\paragraph{Training Protocol and Stopping.}
We use Adam with batch size 256. For the full training set (1,000 folders) we train for 200 epochs
and scale epochs inversely with dataset size (e.g., 400 epochs for 500 folders) to keep the number
of parameter updates approximately fixed. Each epoch we evaluate MI on the full test set and on one
training batch; $d_{\rm eff}$ is computed from that training batch. Because this is a high-MI
regime where InfoNCE approaches its ceiling $\log_2(\text{batch size})\approx 8$ bits, the standard
max-test rule can select $t^\ast$ deep in the saturation region
(Fig.~\ref{fig:supp_pendulum}A). We therefore choose $t^\ast$ as the earliest epoch at which the
test MI reaches a fixed fraction of its maximum. Figure~\ref{fig:supp_pendulum}B shows that
$d_{\rm eff}$ is stable across a broad range of such fractions, and
Fig.~\ref{fig:supp_pendulum_threshs} shows that the qualitative trends in $d_{\rm eff}$ persist
across thresholds and dataset sizes.

\paragraph{Potential Improvements.}
We emphasize that the recovery of physical degrees of freedom reported here was obtained with the
same generic MLP backbone used in the synthetic benchmarks, demonstrating that the method works
out-of-the-box. That said, performance in other high-dimensional modalities will likely benefit
from more tailored architectures (e.g., CNN encoders to exploit spatial structure in images). In
high-information regimes where InfoNCE approaches its $\log_2(\text{batch size})$ ceiling, larger
batch sizes or alternative variational bounds could further reduce saturation effects. We leave
such domain-specific optimizations to future work.

\end{document}